\documentclass[letterpaper,journal]{IEEEtran}

\usepackage{gensymb}
\usepackage{subfigure}
\usepackage{amsmath,amsfonts}
\usepackage{algorithmic}
\usepackage{array}
\usepackage[caption=false,font=normalsize,labelfont=sf,textfont=sf]{subfig}
\usepackage{textcomp}
\usepackage{stfloats}
\usepackage{url}
\usepackage{verbatim}
\usepackage{graphicx}
\usepackage{cite}
\usepackage{color}
\usepackage{booktabs}
\usepackage{threeparttable}
\usepackage{bbding}
\usepackage{pifont}
\usepackage[ruled,linesnumbered]{algorithm2e}
\usepackage{multicol}

\def\BibTeX{{\rm B\kern-.05em{\sc i\kern-.025em b}\kern-.08em
    T\kern-.1667em\lower.7ex\hbox{E}\kern-.125emX}}
\usepackage{balance}

\ifodd 1
\newcommand{\com}[1]{\textbf{\color{red}(COMMENT: #1)}} 
\else

\newcommand{\com}[1]{}
\fi

\newcommand{\revise}[1]{{\color{black}{#1}}}
\newcommand{\TMCrevise}[1]{{\color{black}{#1}}}
\newcommand{\RR}[1]{{\color{black}{#1}}}

\def\fig{Fig.}

\def\eg{e.g.}
\def\ie{i.e.}
\def\aka{a.k.a.}
\def\eqn{Eqn.}

\begin{document}

\newcommand{\mycustomsize}{\fontsize{22}{\baselineskip}\selectfont}
\title{mmE-Loc: Facilitating Accurate Drone Landing \\with Ultra-High-Frequency Localization}

\author{

Haoyang Wang, 
Jingao Xu,
Xinyu Luo,
Ting Zhang,
Xuecheng Chen,
Ruiyang Duan, \\
Yunhao Liu, ~\IEEEmembership{Fellow, ~IEEE},
Jianfeng Zheng,
Weijie Hong,
Xiaoqiang Ji,
Yuqing Tang,
Xinlei Chen

\thanks{


A preliminary version of this work appeared in the 23rd ACM Conference on Embedded Networked Sensor Systems (ACM SenSys 2025) \cite{wang2025ultra}.

Haoyang Wang, Xinyu Luo, Xuecheng Chen, Ting Zhang and Xinlei Chen are with Shenzhen International Graduate School, Tsinghua University, China. (E-mail: \{haoyang-22, luo-xy23, chenxc21\}@mails.tsinghua.edu.cn, zhangt2112@gmail.com, chen.xinlei@sz.tsinghua.edu.cn).

Jingao Xu is with Carnegie Mellon University, USA. (E-mail: jingaox@andrew.cmu.edu)

Ruiyang Duan is with Meituan Academy of Robotics Shenzhen and Meituan Inc., China. (E-mail: duanruiyang@meituan.com)

Yunhao Liu is with the School of Software and BNRist, Tsinghua University, China. (E-mail: yunhao@greenorbs.com)

Jianfeng Zheng and Weijie Hong are with Shenzhen Smart City Communication Co., Ltd., China.  (E-mail: \{hongweijie, zhengjianfeng\}@smartcitysz.com)

Xiaoqiang Ji is with the Chinese University of Hong Kong, Shenzhen. (E-mail: jixiaoqiang@cuhk.edu.cn)

Yuqing Tang is with the International Digital Economy Academy. (E-mail: tangyuqing@idea.edu.cn) 

Corresponding author: Xinlei Chen.

Manuscript submitted: November 2025.
}








\vspace{-0.4cm}
}

\markboth{IEEE TRANSACTIONS ON MOBILE COMPUTING}{}

\maketitle

\begin{abstract}
For precise, efficient, and safe drone landings, ground platforms should real-time, accurately locate descending drones and guide them to designated spots.
While mmWave sensing combined with cameras improves localization accuracy, lower sampling frequency of traditional frame cameras compared to mmWave radar creates bottlenecks in system throughput. 
In this work, we upgrade traditional frame camera with event camera, a novel sensor that harmonizes in sampling frequency with mmWave radar within ground platform setup, and introduce mmE-Loc, a high-precision, low-latency ground localization system designed for precise drone landings.
To fully exploit the \textit{temporal consistency} and \textit{spatial complementarity} between these two modalities, we propose two innovative modules: \textit{(i)} the Consistency-instructed Collaborative Tracking module, which further leverages the drone's physical knowledge of periodic micro-motions and structure for accurate measurements extraction, and \textit{(ii)} the Graph-informed Adaptive Joint Optimization module, which integrates drone motion information for efficient sensor fusion and drone localization.
\RR{
Extensive experiments (30+ hours) demonstrate that mmE-Loc attains 0.083$m$ localization accuracy and 5.12$ms$ end-to-end latency, outperforming four state-of-the-art methods by over 48\% and 62\%, respectively.
}
\end{abstract}




\vspace{-0.5cm}
\section{Introduction}

\begin{figure}[t]
    \centering
        \includegraphics[width=1\columnwidth]{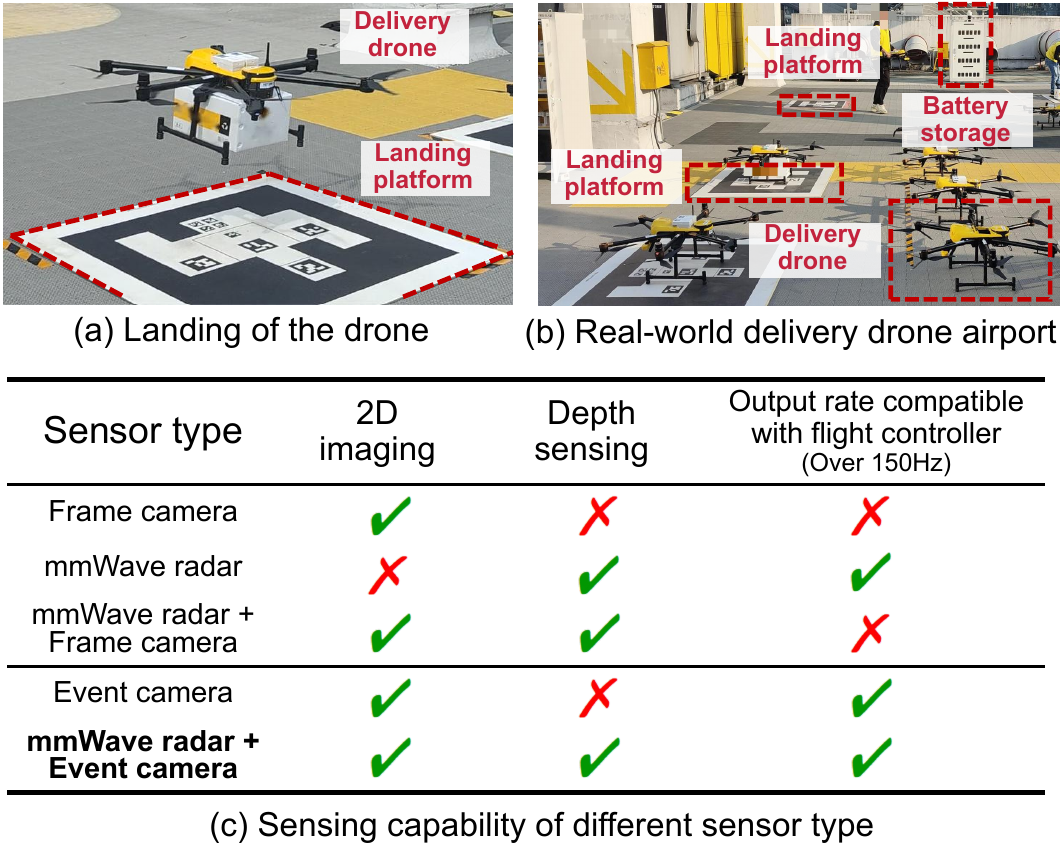}
        \vspace{-0.75cm}
    \caption{
    \TMCrevise{Snapshot of drone landing phase, deliver drone airport, and performance of different sensors.
    (a) A delivery drone lands on the landing platform. 
    (b) The real-world drone airport is equipped with multiple drones for package delivery. 
    (c) Integrating mmWave radar with event camera combines reliable depth sensing and 2D imaging at ultra-high sampling frequencies. 
    This solution achieves high spatio-temporal resolution 2D sensing and precise depth sensing for drone ground localization, while maintaining full compatibility with flight controllers by operating at update frequencies exceeding 150Hz.
    }
    }
    \label{intro}
    \vspace{-0.5cm}
\end{figure}

Projected to soar to a \$1 trillion market by 2040 \cite{low_altitude_eco}, the drone-driven low-altitude economy is transforming sectors with revolutionary applications such as on-demand delivery \cite{chen2024ddl}, meticulous industrial inspections \cite{xu2022swarmmap}, and rapid relief-and-rescue \cite{chen2024soscheduler, zhang2023rf}. 
Of paramount importance within this burgeoning sector is the \textit{landing phase}, where ground platforms locate drones descending from below 10 meters and guide them to accurately land at designated spots (\fig\ref{intro}a) \cite{weiguo2023micnest, sun2022aim}.
\TMCrevise{Situated near populated and commercial zones, these operations emphasize safety and reliability: our research with a leading drone delivery company reveals that a landing bias of just 10$cm$ will result in drones missing charging ports, damaging delivery targets, or even posing risks to people \cite{gonzalez2021visual}. 
Such inaccuracies undermine operational efficiency of this rapidly growing sector, potentially leading to severe consequences.}

Widely adopted and straightforward approaches involve installing cameras at the center or edges of drone landing pads and employing computer vision algorithms for drone localization \cite{li2018real, zhang2019eye}.
However, traditional frame cameras' Achilles heel is capturing only 2D images without depth information, leading to scale uncertainty that limits the 3D localization accuracy \cite{xie2023mozart}.
To address this shortcoming, current practices have incornporated mmWave sensing to provide the lacked depth information for better localization accuracy and reliability in various conditions \cite{deng2022geryon, lu2020milliego}.

\RR{
Albeit inspiring, our benchmark study with a world-leading drone delivery company in landing scenarios (\fig \ref{intro}b) reveals another critical drawback (\fig \ref{intro}c): the long exposure times of frame cameras ($>20ms$) prevent their sampling rates from matching the high frequency of mmWave radars (\eg, $200Hz$). 
This limitation creates system efficiency and throughput bottlenecks, restricting drone location updates to $<50Hz$.
In contrast, drone flight controllers typically require location input rates $>150Hz$ to precisely adjust the drone's flight attitude for safe landing \cite{8412592, sie2023batmobility}.
Therefore, most of the time, drone flight controllers can only adjust the flight attitude based on uncalibrated coarse location updates from mmWave radar or onboard IMU.
However, during the landing phase, precise attitude control is crucial. 
Such coarse updates may lead to a high failure rate in landing, potentially resulting in serious accidents or injuries.
The inefficiency of the system originates from the inherent physical limitations of conventional frame cameras and cannot be easily solved by software solutions.
}

\noindent \textbf{Upgrade frame camera to event camera.}
Event cameras are bio-inspired sensors that report pixel-wise intensity changes with $ms$-level resolution \cite{wang2025towards, gallego2020event}, capturing high-speed motions without blurring \cite{he2024microsaccade, ruan2025pre}, ideal for fast-tracking tasks \cite{luo2024eventtracker}.
Event cameras offer $ms$-level sampling latency, which harmonizes exceptionally with the high sampling frequencies of mmWave radar.
Their 2D imaging capability also complements radars' limited spatial resolution, similar to how traditional frame cameras operate.
\TMCrevise{Such \textit{temporal-consistency} and \textit{spatial-complementarity} across event camera and mmWave radar inspire us to upgrade the frame camera with an event camera, pairing it with the mmWave radar to enable accurate and low-latency drone localization.}


\begin{figure*}[t]
    \centering
        \includegraphics[width=2\columnwidth]{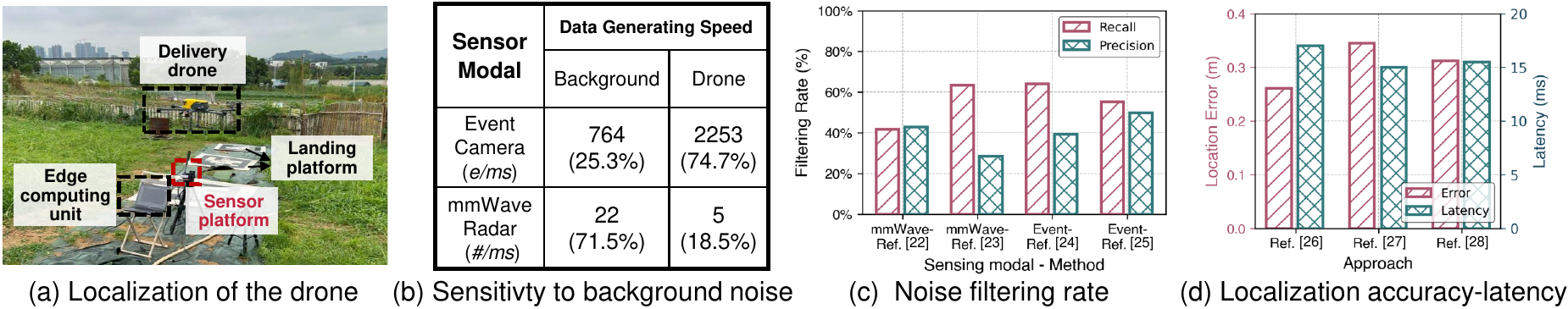}
        \vspace{-0.3cm}
    \caption{
    \TMCrevise{
    Benchmark study on drone localization.
    (a) We conduct the benchmark study at a real-world drone delivery airport.
    \RR{
    (b) Both the event camera and mmWave radar are sensitive to environmental variations, leading to significant sensing noise. Within a \(1~\text{ms}\) interval, event camera background produces \(764\) noise events (\(25.3\%\) of total), while mmWave radar background yields \(22\) noise points (\(71.5\%\) of total). 
    }
    \RR{
    (c) Comparison of noise filtering performance in terms of Recall and Precision. 
    The filtering rate is defined as 
    $\text{Recall} = \frac{N_{\text{filtered, true}}}{N_{\text{noisy}}}$ and 
    $\text{Precision} = \frac{N_{\text{filtered, true}}}{N_{\text{filtered, total}}}$, 
    where $N_{\text{filtered, true}}$ is the number of correctly removed noise samples. 
    A value of $100\%$ indicates the ideal case where all noise samples are correctly removed (Recall = 100\%) 
    and no valid samples are mistakenly removed (Precision = 100\%). 
    Existing algorithms rely solely on either mmWave radar or event cameras, failing to exploit their cross-modal correlations, which limits their noise filtering capability.
    }
    (d) These algorithms are prone to cumulative drift errors and experience considerable delays.
    }
    }
    \label{relatedwork}
    \vspace{-0.3cm}
\end{figure*} 

\noindent \textbf{Our work.}
Following the above insight, we present \textbf{mmE-Loc}, the first active, high-precision, and low-latency landing drone ground localization system that enhances mmWave radar functionality with event cameras. 
mmE-Loc operates effectively in scenarios where the urban canyon effect degrades the accuracy of GPS or RTK systems, particularly as altitude decreases, rendering them nearly ineffective during the landing phase.
With mmE-Loc, drones can achieve reliable localization in such scenarios, even under challenging conditions (e.g., low illumination), ensuring stable and efficient landings.

However, our benchmark study at a real-world drone delivery airport (\fig \ref{relatedwork}a) highlights several challenges that have been solved in making \textbf{mmE-Loc} a viable system outdoors:\\
$(i)$ \textit{How to accurately extract drone-related measurements} given the immense noisy output of event cameras and mmWave radars, which also lack inherent drone semantic information and differ greatly in dimension and pattern?
Both modalities are sensitive to environmental variations, as shown in \fig \ref{relatedwork}b.
Existing algorithms \cite{cao2024virteach, liu2024pmtrack, wang2021asynchronous, alzugaray2018asynchronous} are typically designed for single-modality, resulting in low noise filtering rates (recall and precision $<$ 65\% in \fig \ref{relatedwork}c).\\
$(ii)$ \textit{How to efficiently fuse event camera and mmWave measurements} that are heterogeneous in measurements precision, scale, and density? 
Existing EKF (extended Kalman filter) or PF (particle filter) based methods \cite{falanga2020dynamic,zhao20213d, mitrokhin2018event}, suffer from the cumulative drift errors, making them insufficient for precise drone localization (\fig \ref{relatedwork}d).\\
$(iii)$ \textit{How to further optimize the efficiency of the fusion algorithm} to achieve high-frequency drone ground localization, given the limited computational resources on the landing platforms?
Existing methods experience significant processing delays, rendering them unsuitable for low-latency drone localization tasks (\fig \ref{relatedwork}d) \cite{zhao20213d, falanga2020dynamic, mitrokhin2018event}.

To solve the above challenges, the design of mmE-Loc excel in the three aspects of drone ground localization:

\TMCrevise{\noindent $\bullet$ \textit{On system architecture front.}
\TMCrevise{Upgrading frame camera to event camera with $ms$-level latency to pair with the mmWave radar, mmE-Loc improves accuracy and efficiency of drone ground localization at data source level.}
The system architecture tightly integrates both modalities, from early-stage noise filtering and drone detection to later-stage fusion and optimization, fully leveraging advantages of both sensors (§\ref{3.2}).\\
\noindent $\bullet$ \textit{On system algorithm front.}
We introduce a Consistency-instructed Collaborative Tracking (\textit{CCT}) module, which leverages the drone's physical knowledge of periodic micro-motions and structure, along with cross-modal \textit{temporal-consistency}, to filter environment-induced noise and achieve accurate drone detection (§\ref{4.1}). 
We then present a Graph-informed Adaptive Joint Optimization (\textit{GAJO}) module, which fuses \textit{spatial-complementarity} with a novel factor graph to enhance accuracy of drone ground localization, while incorporating drone motion information as a constraint to improve efficiency, resulting in a trajectory with minimal bias and low cumulative drift (§\ref{4.2}).} \\
\noindent $\bullet$ \textit{On system implementation front.}
\TMCrevise{We further analyze the sources of latency and propose an incremental optimization method to further improve efficiency of the \textit{GAJO} algorithm.}
This approach allows \textit{GAJO} to dynamically optimize a set of locations, maintaining accuracy while reducing latency (§\ref{I}).

\begin{figure*}[t]
    \centering
        \includegraphics[width=2\columnwidth]{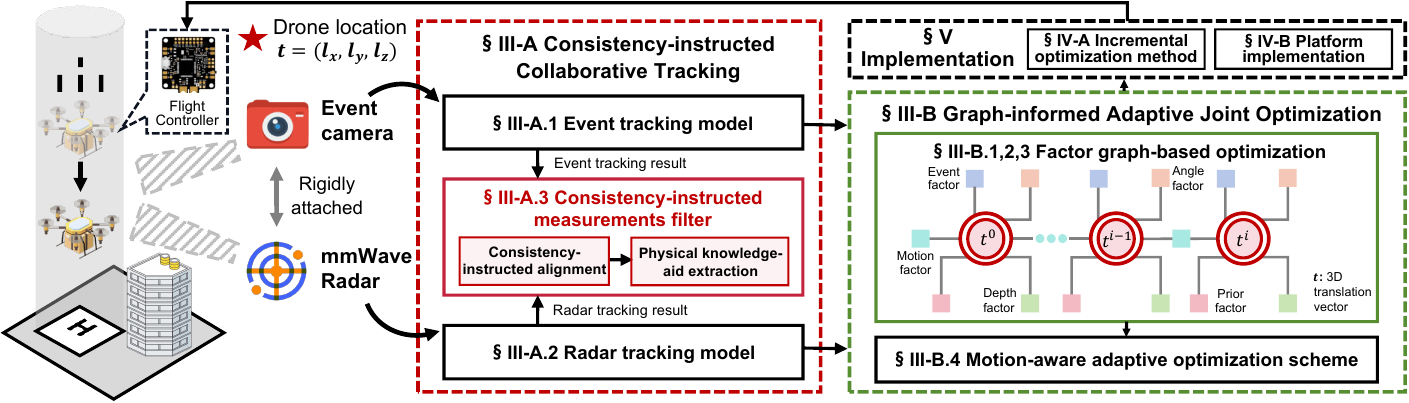}
        \vspace{-0.25cm}
    \caption{\TMCrevise{System architecture of mmE-Loc.
    mmE-Loc consists of two main components. The first is Consistency-instructed Collaborative Tracking, which performs noise filtering on the data output by both the event camera and mmWave radar. It leverages the drone's physical knowledge of periodic micro-motions and structure for drone detection and provides a preliminary localization result for the drone.
    The second component is Graph-informed Adaptive Joint Optimization.
    This module designs various factors such as event factors and depth factors, and incorporates prior knowledge about the drone’s motion through a prior factor. It adapts a factor graph for joint optimization to achieve trajectory optimization for drone. 
    This module also introduces an adaptive window mechanism that dynamically adjusts the joint optimization window based on localization performance, reducing the number of parameters to be optimized.
    For implementation, mmE-Loc further introduces an Incremental Optimization Method to accelerate the solution process.
    }}
    \label{overview}
    \vspace{-0.5cm}
\end{figure*}

We fully implement mmE-Loc using a COTS event camera and mmWave radar. 
Over 30+ hours of indoor and outdoor experiments under various drone flight conditions assess its localization accuracy and end-to-end latency performance against four SOTA methods.
mmE-Loc achieves an average localization accuracy of 0.083$m$ and latency of 5.12$ms$, surpassing baselines by $>$48\% and $>$62\%, respectively, and showing minimal sensitivity to drone type and environment.
We also deploy mmE-Loc at a real-world drone delivery airport (\fig \ref{relatedwork}a) for 10+ hours, demonstrating its practicality for commercial-level drone landing requirements. 

In summary, this paper makes the following contributions.\\
\noindent $(1)$ We explore a novel sensor configuration, event camera plus mmWave radar, that harmonizes ultra-high sampling frequencies and propose mmE-Loc, a ground localization system for drone landings that delivers $cm$ accuracy and $ms$ latency.\\
\noindent $(2)$ \TMCrevise{We present $CCT$, which leverages \textit{temporal consistency} and the drone's physical knowledge of periodic micro-motions and structure for precise drone detection; and $GAJO$, which employs \textit{spatial complementarity} with a novel factor graph, along with drone motion information to enhance localization.} \\
\noindent $(3)$ We implement and extensively evaluate mmE-Loc by comparing it with four SOTA methods, showing its effectiveness. We also deploy mmE-Loc in a real-world drone delivery airport, demonstrating feasibility of mmE-Loc.

\vspace{-0.2cm}
\section{System Overview}
\RR{
mmE-Loc integrates an event camera with mmWave radar for accurate, low-latency drone ground localization, enabling rapid position adjustments for precise landing.
For safety-critical commercial operations, mmE-Loc can complement RTK or visual markers to ensure precise landing.
This section formally defines the problem mmE-Loc addresses and provides an overview of the system design.
}

\vspace{-0.3cm}
\subsection{Problem formulation}


\textbf{Reference systems.} \label{3.2}
There are four reference (\aka, coordinate) systems in mmE-Loc: 
$(i)$ the Event camera reference system $\mathtt{E}$; 
$(ii)$ the Radar reference system $\mathtt{R}$; 
$(iii)$ the Object reference system $\mathtt{O}$;
$(iv)$ the Drone reference system $\mathtt{D}$.
Note that a drone can be considered as an object.
For clarity, before an object is identified as a drone, we utilize $\mathtt{O}$. 
Once confirmed as a drone, we use $\mathtt{D}$ for the drone and continue using $\mathtt{O}$ for other objects.
Throughout the operation of the system, $\mathtt{E}$ and $\mathtt{R}$ remain stationary and are rigidly attached together, while $\mathtt{O}$ and $\mathtt{D}$ undergo changes in accordance with the movement of the object and the drone, respectively. 
The transformation from $\mathtt{R}$ to $\mathtt{E}$ can be readily obtained from calibration \cite{wang2023vital}. 

\textbf{Goal of mmE-Loc.}
The goal of mmE-Loc is to determine 3D location of the drone, defined as $t_{\mathtt{ED}}$, the translation from coordinate system $\mathtt{D}$ to $\mathtt{E}$.
Specifically, mmE-Loc optimizes and reports 3D location of drone $(l_x, l_y, l_z)$ at each timestamp $i$ with input of event stream and radar sample.
$t_{\mathtt{ED}}$ and ($l_x$, $l_y$, $l_z$) are equivalent representations of the drone’s location and can be inter-converted with Rodrigues’ formula \cite{min2021joint}. 
The former representation is adopted in the paper, as it is commonly used in drone flight control systems.

\vspace{-0.3cm}

\subsection{Overview}
As illustrated in \fig \ref{overview}, mmE-Loc comprises two key modules and task-oriented implementation: 

\noindent $\bullet$ 
The \textit{CCT} (\textbf{C}onsistency-instructed \textbf{C}ollaborative \textbf{T}racking) for event camera and mmWave radar noise filtering, drone detection, and preliminary localization of the drone.
This module utilizes time-synchronized event streams and mmWave radar measurements as inputs. 
The \textit{radar tracking model} processes radar measurements to generate a sparse 3D point cloud. 
Meanwhile, the \textit{event tracking model} takes into the stream of asynchronous events for event filtering, drone detection, and tracking. 
\TMCrevise{
Finally, \textit{consistency-instructed measurements filter} aligns the outputs of both tracking models by leveraging \textit{temporal-consistency} between the two modalities. 
It then utilizes the drone's physical knowledge to extract drone-specific measurements and achieve drone preliminary localization. 
Specifically, this method exploits two key characteristics of the landing drone: $(i)$ the high-frequency event bursts caused by the rapid micro-motions of the propellers, and $(ii)$ the drone’s symmetrical structure causes these events to exhibit axial symmetry, both of which serve as distinctive signatures for identifying the drone.

}


\noindent $\bullet$
The \textit{GAJO} (\textbf{G}raph-informed \textbf{A}daptive \textbf{J}oint \textbf{O}ptimization) for fine and low-latency localization and trajectory optimization of the drone.
Based on the operational principles of two sensors and their respective noise distributions, \textit{GAJO} incorporates a meticulously designed \textit{factor graph-based optimization} method. 
\TMCrevise{
This module employs the \textit{spatial-complementarity} from both modalities to unleash the potential of event camera and mmWave radar in drone ground localization.
Specifically, \textit{GAJO} jointly fuses the preliminary location estimation from the \textit{event tracking model} and the \textit{radar tracking model} and refines them.
To reduce latency, we thoroughly analyze its underlying sources and introduce a   motion-aware adaptive optimization scheme. 
By exploiting the drone's inherent motion information, this method triggers optimization at appropriate times and adaptively adjusts the optimization window size, achieving precise localization with $ms$-level latency.

}
\vspace{-0.3cm}
\section{System Design} \label{4}
In this section, we introduce CCT for noise filtering, detection, and preliminary localization of drone (§ \ref{4.1}). 
Subsequently, we delve into GAJO for fine and low-latency localization and trajectory optimization of drone (§ \ref{4.2}).

\vspace{-0.2cm}
\subsection{\textit{CCT}: Consistency-instructed Collaborative Tracking} \label{4.1}

mmWave radar is prone to multipath effects, causing inaccurate point clouds, while event cameras, though capturing per-pixel brightness changes asynchronously, are easily affected by non-drone factors such as shadows.
The lack of intrinsic drone semantics and the large modality gap between radar and event data make noise filtering challenging, creating detection bottlenecks that degrade localization performance.
We therefore focus on improving noise filtering and drone detection to enable reliable preliminary localization.

\RR{
To address this challenge, we examine operational principles of both sensors.
Our design is guided by two key observations:
\textit{(i)Event cameras and mmWave radar exhibit temporal consistency while operating via distinct mechanisms.} 
Both maintain $ms$-level latency, with event cameras immune to multipath effects and mmWave radar unaffected by changes in brightness.
\textit{(ii) Drones display periodic micro-motion features (e.g., propeller rotation) and a symmetrical structure}, providing stable and distinctive cues. 
The high-speed rotation of propellers induces rapid light changes, generating numerous events, which exhibit axial symmetry due to drone’s structure}.

\begin{figure}
\setlength{\abovecaptionskip}{-0.1cm} 
  \begin{minipage}[t]{1\columnwidth}
    \centering
    \includegraphics[width=0.7\columnwidth]{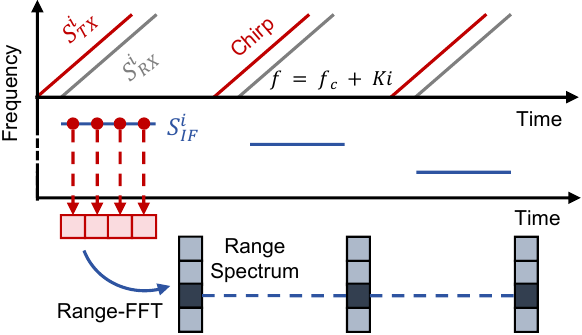}
    \caption{Distance calculation by frequency difference}
    \label{CCT(a)}
  \end{minipage}
  \begin{minipage}[t]{1\columnwidth}
    \centering
    \includegraphics[width=0.7\columnwidth]{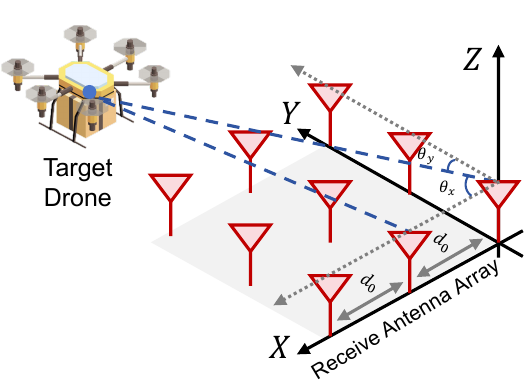}
    \caption{Direction calculation by phase difference}
    \label{CCT(b)}
  \end{minipage}
  \hfill
  \vspace{-0.5cm}
\end{figure}

To realize this idea, we design \textit{CCT}, a lightweight cross-modal drone detector and tracker.
\textit{CCT} includes three components:
$(i)$ a radar tracking model (§\ref{4.1.1}) that provides sparse point clouds encoding the distance and direction of objects;
$(ii)$ an event tracking model (§\ref{4.1.2}) for event filtering, detection, and tracking of objects;
\RR{
$(iii)$ a consistency-instructed measurements filter (§\ref{4.1.3}) that leverages temporal consistency between modalities and the drone’s periodic micro-motion features to extract drone-related events and point clouds, enabling preliminary localization.
}

\subsubsection{\textbf{Radar tracking model}} \label{4.1.1}
In this part, we calculate the distance $D$ and direction vector $\vec{v}$ between the radar and objects, along with a preliminary estimate of object’s location.

\textbf{Distance calculation.} 
As shown in \fig \ref{CCT(a)}, the frequency difference between the transmitted (TX) and received (RX) signals indicates the signal propagation time, revealing distance between object and radar.
Denoting $D^i$ as the distance at time $i$, TX and RX signals as:
\begin{equation}
\begin{aligned}
S_{TX}^i\!=\!\exp \left[j\left(2 \pi f_c i+\pi K i^2\right)\right], 
\TMCrevise{S_{RX}^i\!=\!\alpha S_{TX}\left[i-\frac{2D^i}{c}\right],}
\label{TX_RX}
\end{aligned}
\end{equation}
where $\alpha$ denotes the attenuation rate, $f_c$ is the initial frequency, $K$ represents the chirp slope of FMCW signal, and $c$ stands for light speed.
The TX and RX signals from \eqn (\ref{TX_RX}) undergo mixing and a low-pass filter (LPF) to extract the intermediate frequency signal (IF signal) $s(t)$: 
\begin{equation}
S_{IF}^i=LPF(S_{TX}^{i*} S_{RX}^{i}) 
\TMCrevise{\approx \alpha \exp \left[j 2 \pi\left(\frac{2KD^i}{c}\right)i\right]}.
\end{equation}
The frequency value $f_{IF}$ within $S_{IF}^i$ encapsulates distance information. 
After the Range-FFT operation of $S_{IF}^i$, $f_{IF}$ is extracted, facilitating distance calculation as
\begin{equation}
D^i=\frac{c f_{IF}}{2K}.
\label{distance}
\end{equation}

\textbf{Direction calculation.}
Using a fixed antenna array, the mmWave radar determines the object's direction by employing two orthogonal linear arrays. 
As depicted in \fig \ref{CCT(b)}, each linear array captures an Angle of Arrival (AoA), calculated from the phase difference between adjacent antennas spaced apart by $d$ as $cos \theta = \Delta \phi \lambda/2 \pi d$, where $\theta$ represents AoA, $\lambda$ denotes the wavelength and $\Delta \phi$ indicates the phase difference. 
With two orthogonal arrays, the radar obtains two AoAs, $\theta_x$ and $\theta_y$. The unit vector indicating object's direction $i$ is given by
\begin{equation}
\vec{v}^i=[\cos \theta_x \cos \theta_y \sqrt{1-\cos ^2 \theta_x-\cos ^2 \theta_y}]^{\mathrm{T}}.
\label{angle}
\end{equation}

Using the distance and angle information obtained above, along with the spatial relationship between radar and event camera, we can determine the preliminary 3D location estimation of object in $\mathtt{E}$ as $P_E = D\vec{v}+t_{ER}$.
\RR{We then leverage the mmWave radar for object 3D location tracking, estimating the translation $t_{\mathtt{EO}}$ of object at time $i$:
\begin{equation}
\begin{aligned}
\vspace{-0.2cm}
t_{\mathtt{EO}}^i =t_{\mathtt{EO}}^{i-1}+U_{\mathtt{E}}^{i}+w^i + w^{i-1}, 
U_{\mathtt{E}}^{i} = \left(P_{\mathtt{E}}^i-P_{\mathtt{E}}^{i-1}\right)
\vspace{-0.2cm}
\label{translation}
\end{aligned}
\end{equation}
where $w_i$ and $w_{i-1}$ signify measurement noise.
}

\RR{
\textbf{Drawback of mmWave Radar.}
Although mmWave radars are effective for estimating object depth along the radial direction, their accuracy degrades when capturing motion in the horizontal and vertical (tangential) directions \cite{zhang2023push}.
To address this limitation, we integrate an event camera, which features comparable latency but relies on a fundamentally different sensing mechanism.
With its high spatial and temporal resolution, the event camera enhances object detection and provides precise motion information in the tangential plane, effectively complementing the mmWave radar.
}

\begin{figure}
\setlength{\abovecaptionskip}{-0.1cm} 
  \begin{minipage}[t]{1\columnwidth}
    \centering
    \includegraphics[width=0.6\columnwidth]{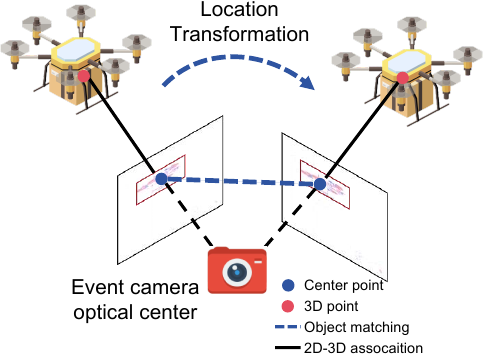}
    \caption{Event tracking model}
    \label{CCT(c)}
  \end{minipage}
  \begin{minipage}[t]{1\columnwidth}
    \centering
    \includegraphics[width=0.95\columnwidth]{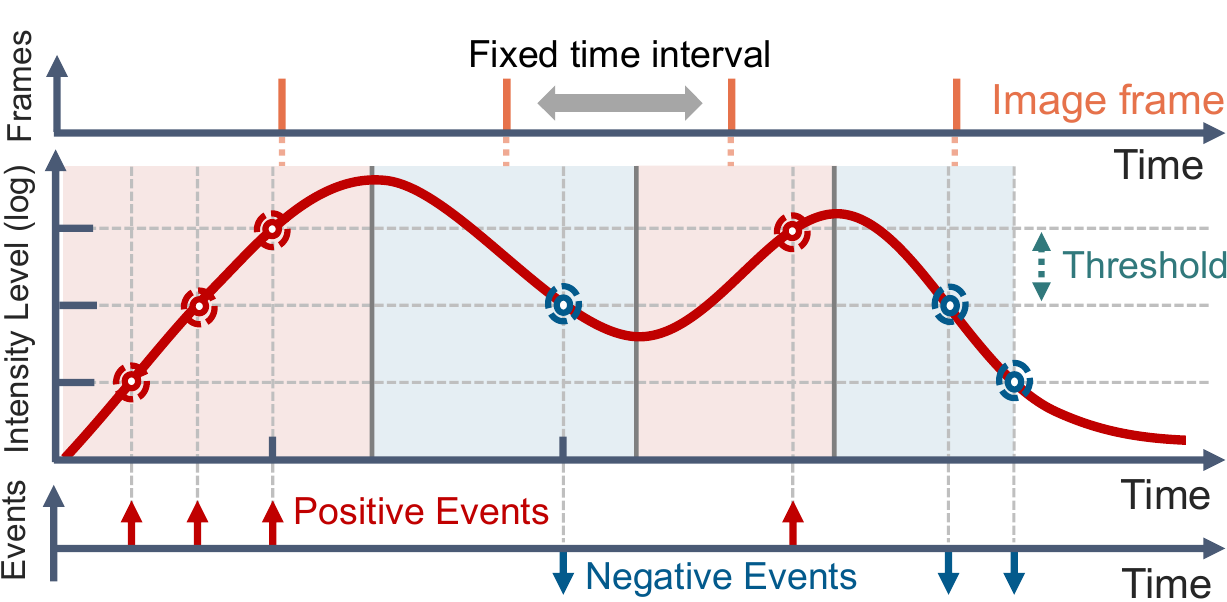}
    \caption{
    \RR{
    Illustration of synchronous images and asynchronous events. 
    Frame cameras capture images at fixed intervals using a global shutter. 
    In contrast, each pixel in an event camera responds independently, generating events whenever intensity changes exceed a threshold.
    }
    }
    \label{event}
  \end{minipage}
  \hfill
  \vspace{-0.5cm}
\end{figure}

\subsubsection{\textbf{Event tracking model}} \label{4.1.2}
In this part, we demonstrate the process of noise filtering from a stream of asynchronous events, and how to detect and track objects with the filtered events, as depicted in \fig \ref{CCT(c)}.
\RR{
As illustrated in \fig \ref{event}, unlike conventional frame-based cameras that rely on a global shutter to capture images at fixed intervals, event cameras asynchronously record per-pixel intensity changes with $ms$-level temporal resolution. 
This enables high-speed motion capture without motion blur, but also introduces challenges in noise suppression and object detection.
}

\begin{figure}
\setlength{\abovecaptionskip}{-0.1cm} 
  \begin{minipage}[t]{1\columnwidth}
    \centering
    \includegraphics[width=0.95\columnwidth]{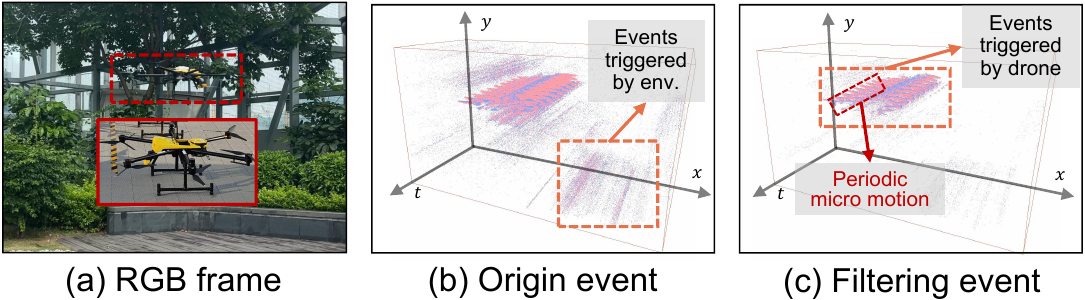}
    \caption{
    Event filtering performance. 
    \RR{
    mmE-Loc filters event noise via neighborhood similarity and manages events with SAEs, ensuring precise spatio-temporal representation while reducing event load and computation.
    }
    }
    \label{performance(a)}
  \end{minipage}
  \begin{minipage}[t]{1\columnwidth}
    \centering
    \includegraphics[width=0.95\columnwidth]{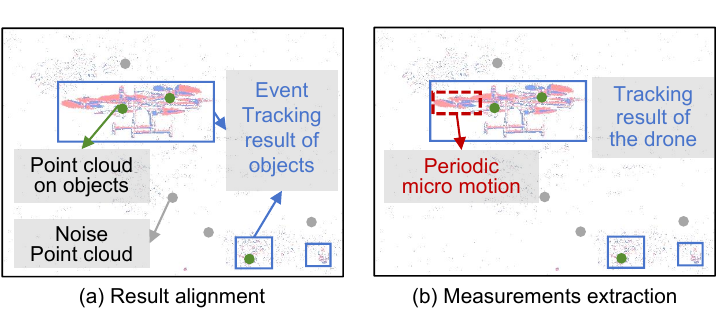}
    \caption{
    Consistency-instructed Measurements Filter.
    \RR{
    (a) Leveraging the temporal consistency of event camera and mmWave radar, we align the outputs of two sensors, enabling the detection of the drone and other objects.
    (b) The rotation of the drone’s propellers induces rapid light intensity changes, generating numerous events with elliptical patterns.
    The drone’s symmetrical structure yields axially symmetric event patterns, providing a key feature for distinguishing it from other objects.
    }
    }
    \label{performance(b)}
  \end{minipage}
  \hfill
  \vspace{-0.5cm}
\end{figure}

\textbf{Similarity-informed event filtering.}
Event cameras are prone to noise from transistor circuits and other non-idealities, requiring pre-processing filtering. 
For the $i^{th}$ event $e^i_{(x, y)}$ with the timestamp $t^i_{(x, y)}$, we assess the timestamp ($t^i_{n(x, y)}$) of the most recent neighboring event in all directions. 
Events with a time difference less than the threshold $T_n$ are retained, indicating object activity, while those exceeding it are discarded as noise (\fig \ref{performance(a)}b, \fig \ref{performance(a)}c).
\revise{
We utilize the Surface of Active Events (SAE) \cite{lin2020efficient} to manage events, mapping coordinates $(x, y)$ to timestamps $(t_l, t_r)$.
Upon a new event's arrival, $t_l$ updates accordingly, and $t_r$ updates only if the previous event at the same location occurred outside the time window $T_k$ or had a different polarity. 
Events that update value of $t_r$ are retained.
The event stream, segregated by polarity, is processed with distinct SAEs. 
This method ensures precise spatial-temporal representation, conserving computational resources.
}

\textbf{Filter-based detection.} 
We employ a grid-based method to cluster events to facilitate object detection. 
The camera's field of view is partitioned into elementary cells sized $c_w \times c_h$. 
For each cell, we compare the event count within a specified time interval ($c_{\Delta t}$) to an activation threshold $c_{thres}$. 
Cells surpassing $c_{thres}$ are marked as active and connected to form clusters, serving as object detection results, including those generated by the drone.
\RR{
For tracking, we deploy Kalman filter-based trackers with a constant velocity motion model, which provides low-latency estimates \cite{zha2025dimm}.
Specifically, the tracker predicts the current state of the object and associates it with the input cluster that has the largest Intersection over Union (IoU).
The selected input cluster then updates the tracker state, resulting in bounding box estimation of moving objects.
}

\RR{
Using bounding box proposals and a pinhole camera model with projection function $\pi$, we estimate preliminary 3D locations of objects.
Specifically, $\pi$ transforms a 3D point $\textbf{X}_\mathtt{E}\!=\![X_\mathtt{E}, Y_\mathtt{E}, Z_\mathtt{E} ]^T$ into a 2D pixel $x$ in image plane as: 
\begin{equation}
\begin{aligned}
x\!=\!\pi\left(\textbf{X}_\mathtt{E}\right)\!&=\![f_x \frac{X_\mathtt{E}} {Z_\mathtt{E}}+c_x,
f_y \frac{Y_\mathtt{E}}{Z_\mathtt{E}}+c_y]^T, 
\end{aligned}
\label{projection_function}
\end{equation}
where $[f_x, f_y]^T$ is focal length of event camera, and $[c_x, c_y]^T$ denotes principal point, both being intrinsic parameters. 
}

Then, the object's preliminary location is estimated using the center point of the bounding box $x^i$ and \eqn (\ref{projection_function}) as: 
\begin{equation}
x^i =\pi(\textbf{X}_\mathtt{E}^i)+v^i =\pi(\textbf{X}_\mathtt{O}^i+t_{\mathtt{EO}}^i)+v^i,
\label{event_tracking_model}
\end{equation}
where $\textbf{X}_\mathtt{O}^i$ represents the corresponding 3D point of center point $x^i$ in the object reference $\mathtt{O}$, $v^i$ denotes the random noise.

\subsubsection{\textbf{Consistency-instructed measurements filter}} \label{4.1.3}

\RR{
The \textit{event tracking model} detects drones and other objects causing light variations, such as indicator lights or shadows, and needs to distinguish drone-induced events from background noise.
Similarly, the \textit{radar tracking model} produces a 3D point cloud containing both valid reflections and multipath artifacts, requiring the extraction of drone-specific points.
}

\textbf{Consistency-instructed alignment.} 
Utilizing the \textit{temporal-consistency} from the event camera and mmWave radar, and their distinct mechanisms respond to dynamic objects, we can filter events affected by lighting variations on stationary objects and vice versa for radar points influenced by multipath effects.
\RR{
Specifically, we align temporal-synchronized radar points to event bounding boxes using spatial relationships of event camera and mmWave radar (\fig \ref{performance(b)}a). 
Using the event camera's projection function \eqn (\ref{projection_function}), we determine that the object's location lies along the ray from the optical center through the bounding box center. 
The system then identifies the nearest radar points along this ray to isolate the object-associated points.
If no point is found, the bounding box is considered noise and discarded.
}

\begin{figure}[t]
    \centering
        \includegraphics[width=0.7\columnwidth]{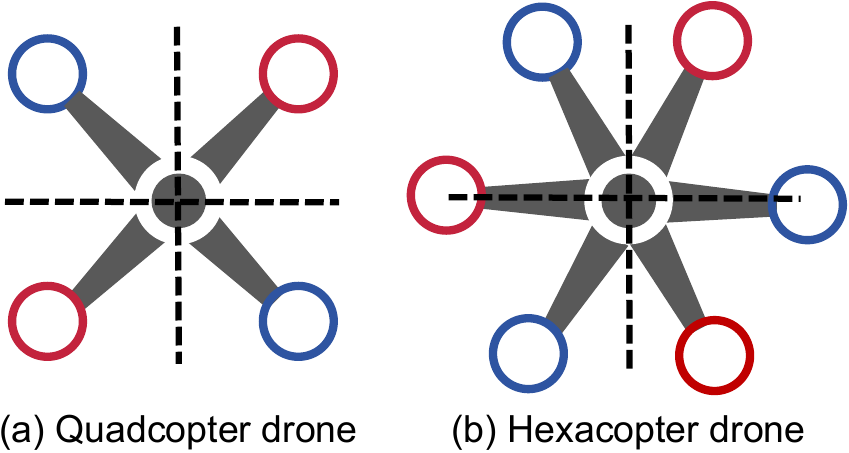}
        \vspace{-0.3cm}
    \caption{
    \TMCrevise{
    The structure of both the quadcopter and hexacopter drones features propellers arranged in axial symmetry.}
    }
    \label{structure}
    \vspace{-0.5cm}
\end{figure} 

\textbf{Drone physical knowledge-aid extraction.} 
\RR{
Since each platform supports one drone landing at a time, we need to identify features of the landing drone and use them to extract landing drone-specific measurements from the aligned tracking results. 
Our findings reveal three distinctive features for landing drones: \textit{(i)} periodic micro-motions (e.g., propeller rotation) observed as temporal oscillations in the event stream; \textit{(ii)} structural event patterns  (e.g., circular patterns) produced by those micro-motions; and \textit{(iii)} symmetry-induced features (e.g., axial symmetry) arising from drone’s multi-propeller design.
Together, these complementary features provide cues for reliably detecting landing drone from surrounding objects.}

$(i)$ \textit{Periodic micro motion-based extraction.}
\RR{
The rapid periodic micro-motions of the propellers produce high-frequency bursts of events, leading to a substantially larger event count for drones compared to surrounding objects. Further details are provided in Appendix A.
Leveraging this property, we transform the spatio-temporal distribution of events into a heatmap and apply statistical metrics to isolate drone-induced measurements.
Specifically, within each time window $[i, i + \delta i]$, events are aggregated into a 2D histogram, where each bin corresponds to a spatial region (e.g., $5\times 5$pixels).
Bins containing propeller rotations accumulate more events due to rapid light intensity changes.
Moreover, propeller motion produces bipolar events within a bin, while background motion and noise typically generate unipolar events (e.g., flying birds).
We therefore identify propeller-related bins by considering both event count and positive-to-negative event ratio, selecting bins with higher counts and a more balanced polarity distribution.
}

\begin{figure}[t]
    \centering
        \includegraphics[width=0.85\columnwidth]{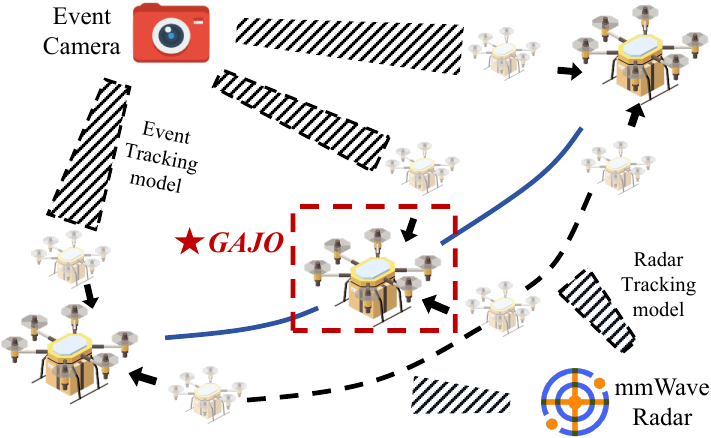}
        \vspace{-0.3cm}
    \caption{
    \TMCrevise{
    Illustration of relationship between \textit{GAJO}, event tracking model and radar tracking model. The \textit{GAJO} module harness \textit{spatial-complementarity} of both modalities through a factor graph-based join optimization, where event camera provides 2D imaging and mmWave radar delivers depth information.}
    }
    \label{relationship}
    \vspace{-0.5cm}
\end{figure} 

$(ii)$ \textit{Drone structure-based extraction.}
\RR{
To further suppress irrelevant measurements, we exploit both the distinct structural patterns of propeller-induced events and the inherent symmetry of the drone’s propeller configuration.}
Specifically, for selected bins, we analyze their connectivity to form clusters and fit ellipses to structure of the resulting clusters. 
\RR{
Given the drone’s four- or six-propeller axial symmetry (\fig \ref{structure}), we evaluate the symmetry of ellipse-fitted clusters to determine whether a bounding box originates from the drone.
Bounding boxes with axially symmetric clusters are retained as drone-generated.}

\RR{
In this way, we associate the event tracking results with their corresponding point clouds, and extract drone tracking outputs from two models for preliminary localization, as shown in \fig \ref{performance(b)}b.
When multiple drones are scheduled to land, they descend and land sequentially. 
This method identifies landing drone and extracts relevant measurements.
}

\vspace{-0.1cm}
\subsection{\textit{GAJO}: Graph-informed Adaptive Joint Optimization} \label{4.2}

The preliminary drone location estimations from the event and radar tracking models suffer from biases. 
Specifically, estimations from event tracking model face scale uncertainty, while estimations from radar tracking model struggle with limited spatial resolution, scatter center drift, and accumulating drift. 
Additionally, estimations from different models are heterogeneous in precision, scale, and density, complicating the fusion and optimization.
Therefore, in this part, we prioritize accurate drone ground localization and trajectory tracking.

\RR{
Our design is based on the insight that \textit{the event and radar tracking models provide spatially complementary features}.
The 2D imaging capability of the event camera and the depth sensing of the mmWave radar can effectively compensate for each other, as illustrated in \fig \ref{relationship}.
Because both the event stream and mmWave samples are drone-related, fully exploiting their \textit{spatial complementarity} through joint optimization can enhance localization performance, leading to trajectories with reduced bias and minimized cumulative drift.
}

\RR{
To realize this idea, we propose \textit{GAJO}, a factor graph-based framework for low-latency and accurate 3D drone localization (§\ref{4.2.1}).
\textit{GAJO} consists of two tightly coupled modules: $(i)$ short-term inter-SAE tracking and $(ii)$ long-term local location optimization, which together improve localization precision (§\ref{4.2.3}).
It also incorporates prior knowledge of the drone’s flight dynamics to refine the trajectory.
To reduce computational cost, \textit{GAJO} leverages motion cues to evaluate localization quality, deciding when long-term optimization is needed and adaptively adjusting the optimization window size to minimize overhead.
}

\begin{figure*}[t]
    \centering
        \includegraphics[width=1.95\columnwidth]{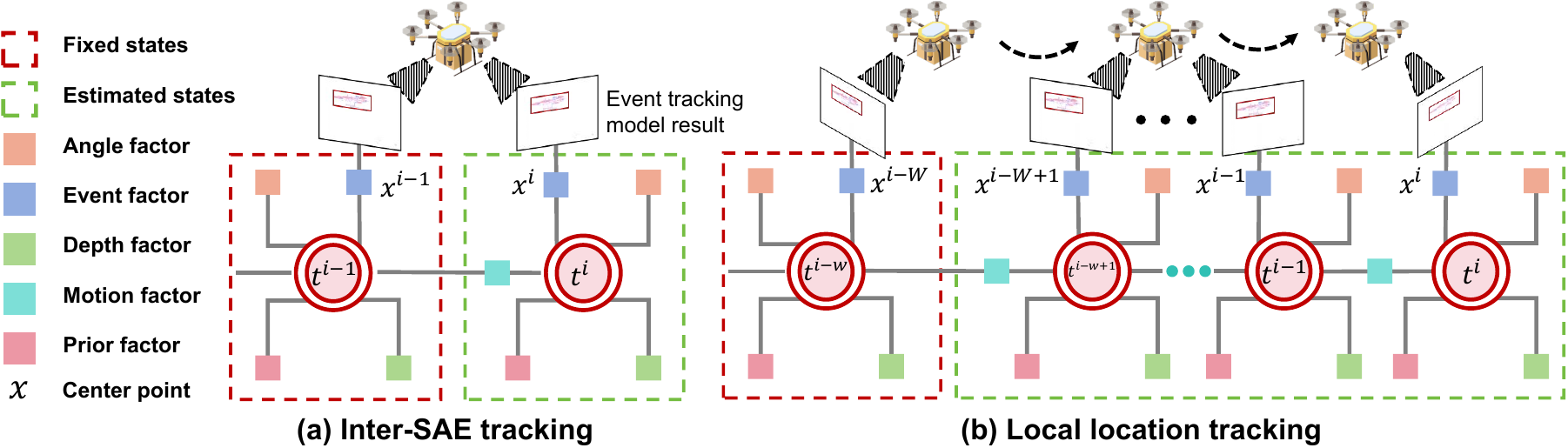}
        \vspace{-0.3cm}
    \caption{
    \TMCrevise{Illustration of Graph-informed Adaptive Joint Optimization. The module consists of two parts. The first part is (a) Inter-SAE Tracking, which aims to perform short-term joint optimization for low-latency output. However, the localization results from this part may accumulate errors over time. Therefore, the second part, (b) Local Location Optimization, performs long-term joint optimization by jointly considering multiple locations and associated constraints, resulting in a trajectory with reduced bias and minimized cumulative drift.}
    }
    \label{factorgraph}
    \vspace{-0.5cm}
\end{figure*}

\subsubsection{\textbf{Factor graph-based optimization}}\label{4.2.1}

A factor graph comprises variable nodes, indicating the states to be optimized (\eg, $t_{ED}^i$), and factor nodes, representing the probability of certain states given a measurement result. 
\RR{
In mmE-Loc, measurements are derived from the Event Tracking (ET) model ($x^i$ from \eqn (\ref{event_tracking_model})) and Radar Tracking (RT) model ($D^i$ from \eqn (\ref{distance}), $\vec{v}^i$ from \eqn (\ref{angle}), and $U_E^{i}$ from \eqn (\ref{translation})).
To estimate the values of a set of variable nodes $\boldsymbol{\mathcal{X}} = \{t_{ED}^i | i \in \mathcal{T}\}$ given measurements $\boldsymbol{\mathcal{Z}} = \{x^i, D^i, \vec{v}^i, U_E^{i} | i \in \mathcal{T}\}$, \textit{GAJO} optimizes connected factors based on maximum a posteriori estimation:
\begin{align}
\begin{split}
\hat{\boldsymbol{\mathcal{X}}} =\underset{\boldsymbol{\mathcal{X}}}{\arg \max } \ 
p(\boldsymbol{\mathcal{X}}) \prod_{i \in \mathcal{T}} \ p\left(x^i, D^i, \vec{v}^i, U_E^{i} \mid t_{ED}^i\right),
\end{split}
\label{factor_graph}
\end{align}
where $p(\boldsymbol{\mathcal{X}})$ represents the prior term, encoding the drone’s flight dynamics and motion constraints.
The $p\left(x^i \mid t_{ED}^i\right)$ corresponds to the likelihood of the ET model, capturing the measurement uncertainty in the SAE plane.
The $p\left(D^i, \vec{v}^i, U_E^{i} \mid t_{ED}^i\right)$ denotes the likelihood of the RT model, which integrates distance, direction, and coarse translation to constrain the drone’s 3D position.
Further details are provided in Appendix B.
}

\subsubsection{\textbf{Probabilistic representation}} \label{4.2.2}
Inferring the drone’s location requires incorporating both the prior and likelihood terms in \eqn \eqref{factor_graph}.
This section provides a formulation of these terms.

\textbf{Prior term.} 
The prior term $p(t_{ED}^i)$ models the location probability distribution of the drone at time $i$ independent of current measurements.
It is derived from a constant velocity motion model, which assumes that the drone maintains a steady speed over short time intervals.
Based on this assumption, the prior location can be predicted as:
\begin{equation}
\vspace{-0.1cm}
\bar{t}_{\mathrm{ED}}^i - t_{\mathrm{ED}}^{i-1} = t_{\mathrm{ED}}^{i-1} - t_{\mathrm{ED}}^{i-2}.
\label{prior_term}
\end{equation}

\textbf{ET model likelihood.} 
The likelihood $p(x^i|t_{ED}^i)$ from the ET model characterizes the distribution of the detected center point given the drone’s location at time $i$.
Following common practice in tracking systems \cite{campos2021orb}, the center point noise $v^i$ is modeled as Gaussian, which has proven effective in capturing localization uncertainty.
Therefore, the likelihood of the ET model is formulated as:
\begin{equation}
p(x^i|t_{ED}^i) \sim \mathcal{N}(\pi(\textbf{X}_E^i), \sigma_{ET}),
\label{ET_model}
\end{equation}
where $\sigma_{ET}$ denotes standard deviation of center point noise.

\textbf{RT model likelihood.}
The likelihood $p(D^i \mid t_{ED}^i)$, $p(\vec{v}^i \mid t_{ED}^i)$, and $p(U_E^{i} \mid t_{ED}^i)$ describe the probability distributions of the measured distance, angle, and motion given the drone’s location at time $i$.
These measurements are modeled as Gaussian distributions:
\begin{equation}
\begin{aligned}
p(D^i \mid t_{ED}^i&) \sim  \mathcal{N}(||t_{ED}^i||, \sigma_{D}), p(\vec{v}^i \mid t_{ED}^i) \sim \mathcal{N}(\vec{v}_{t_{ED}^i}, \sigma_{\vec{v}}), \\
& p(U_E^{i} \mid t_{ED}^i) \sim \mathcal{N}(t_{ED}^i - t_{ED}^{i - 1}, \sigma_{U_E}),
\vspace{-0.4cm}
\label{RT_model}
\end{aligned}
\end{equation}
where $\sigma_{D}$, $\sigma_{\vec{v}}$, and $\sigma_{U_E}$ denote the standard deviations of the distance, angle, and motion measurements, respectively.

\RR{
\subsubsection{\textbf{Fusion-based tracking}} \label{4.2.3}
In mmE-Loc, two fusion schemes are employed for sensor fusion and optimization, as depicted in \fig \ref{factorgraph}.
The first, short-term inter-SAE tracking, provides instantaneous location estimates by minimizing errors across different tracking models.
The second, long-term local location optimization, improves overall trajectory accuracy through joint optimization of a selected set of locations.

\textbf{Short-term inter-SAE tracking.}
Once the measurements of ET model and RT model $(x^i, D^i, \vec{v}^i, U_E^i)$ received, the prior factor, ET factor and the RT factor are formulated using \eqn \eqref{prior_term}, \eqn \eqref{ET_model} and \eqn \eqref{RT_model} as follows:
\begin{equation}
\begin{aligned}
E^i_{\text {Prior }} & =-\log p\left(t_{ED}^i\right),
E^i_{\mathrm{ET}} =-\log p\left(x^i \mid t_{ED}^i\right) \\
& E^i_{\mathrm{RT}} =-\log p\left(D^i, \vec{v}^i, U_E^i \mid t_{ED}^i\right).
\label{factor}
\end{aligned}
\end{equation}

Building on the factors in \eqn \eqref{factor}, the inter-SAE tracking in \fig \ref{factorgraph}a is performed to provide an instantaneous location estimate according to \eqn \eqref{factor_graph} as follows:
\begin{equation}
\begin{aligned}
\hat{t}_{ED}^i = \underset{\boldsymbol{t_{ED}^i}}{\arg \min } \left( E^i_{\text {prior }} + E^i_{\mathrm{ET}} + E^i_{\mathrm{RT}}\right).
\end{aligned}
\label{inter_frame}
\end{equation}
Further details and derivations are provided in Appendix C.

\textbf{Local location optimization.}
To effectively mitigate cumulative localization drift, we perform a local location optimization procedure that refines the estimated positions by exploiting the temporal consistency across multiple consecutive SAEs.
Specifically, this process involves the joint optimization of a set of SAE locations, denoted as $\boldsymbol{\mathcal{X}}=\underset{i \in \mathcal{T}}{\bigcup}\left\{t_{ED}^i\right\}$, as illustrated in \fig \ref{factorgraph}b, where $W = |\mathcal{T}|$ represents the number of SAEs included in the optimization window.
The optimization problem is formulated based on Eq.~\eqref{factor_graph} as follows:
\begin{equation}
\begin{aligned}
\hat{\boldsymbol{\mathcal{X}}} = \underset{\boldsymbol{\mathcal{X}}}{\arg \min } \sum_{i \in \mathcal{T}}\left(E_i^{\mathrm{prior}}+E_i^{\mathrm{ET}}+E_i^{\mathrm{RT}}\right) .
\end{aligned}
\label{local_location}
\end{equation}
Further details and derivations are provided in Appendix D.
It is worth noting that $(i)$ the conditions under which local location optimization is triggered, $(ii)$ the size of the optimization window $\mathcal{T}$ ($W = |\mathcal{T}|$), and $(iii)$ the strategy for solving the inter-SAE tracking and local location optimization problems have a direct impact on both the latency and the accuracy of localization.
To address these issues, we enhance the efficiency of GAJO by introducing a motion-aware adaptive optimization scheme to handle $(i)$ and $(ii)$, and an incremental optimization method to efficiently solve $(iii)$.
}

\RR{
\subsubsection{\textbf{Motion-aware adaptive optimization scheme}}  \label{4.2.4}

In this section, we address two key issues: $(i)$ determining when to trigger the local location optimization, and $(ii)$ defining the size of the optimization window $\mathcal{T}$.
To tackle these problems, we leverage the drone’s onboard motion information. 
Specifically, when mmE-Loc computes the drone’s position, this estimated position is sent to the flight controller. 
Meanwhile, the flight controller typically employs an Inertial Measurement Unit (IMU), which continuously provides the drone’s acceleration measurements.
By integrating the acceleration over time, we can estimate the drone’s displacement between two consecutive timestamps.
Using this motion information, we combine the location estimates from mmE-Loc at time $i-j$ ($j \in \{1, ..., W\}$) to predict the drone’s position at time $i$.
If the predicted position at time $i$ deviates from the mmE-Loc output at the same time by more than a predefined threshold $\Delta$, the local location optimization is triggered for the location set $\boldsymbol{\mathcal{X}}=\underset{k \in [i-j, i]}{\bigcup}\left\{t_{ED}^k\right\}$.

}

\section{Implementation} \label{I}
\subsection{Incremental optimization method} \label{I-1}
\RR{
To efficiently solve the nonlinear least-squares estimation problems, we formulate them within a factor graph framework and employ QR-based matrix factorization to estimate the optimal locations. Nonetheless, the conventional approach of re-linearizing and regenerating the square root information matrix $R$ upon receiving new measurements incurs considerable computational cost, hindering real-time applicability. 
To overcome this limitation, we develop an incremental optimization method that exploits the insight that new measurements predominantly affect localized regions of the factor graph. 
By incrementally updating $R$ for these regions and selectively re-generating it under significant structural changes, the proposed method effectively balances computational efficiency and estimation accuracy. The detailed algorithmic steps are presented in the Appendix E.
}

\subsection{Platform implementation}

\textbf{Sensor platform configuration.}
As illustrated in \fig \ref{setup}, we implement our sensing platform with multiple sensors including 
$(i)$ A Prophesee EVK4 HD evaluation kit, featuring the IMX636ES event-based vision sensor for HD event data (1280 $\times$ 720 pixels) with 47.0\degree  FoV.
$(ii)$ A Texas Instruments (TI) IWR1843 board for transmitting and receiving mmWave signals within the frequency range of 76 $GHz$ to 81 $GHz$ with three transmitting antennas and four receiving antennas.
These antennas are arranged in two linear configurations on the horizontal plane.
$(iii)$ An Intel D435i Depth camera for RGB image capture used in the baseline method.

\TMCrevise{
\textbf{Deployment detail.}
During the experiments, the sensor platform is placed at the center of the landing pad. 
In the real-world case study conducted at an airport, the platform is positioned at the edge of the pad for safety reasons, with the distance from the platform to the pad’s center being less than 1 meter. 
Once the drone takes off, it enters the event camera’s field of view (FoV). All sensors are synchronized via the Robot Operating System (ROS).
mmE-Loc runs on a PC with Ubuntu 20.04, equipped with an Intel i7 CPU, 32GB of RAM, and an NVIDIA GeForce GTX 1070 GPU.
For practical deployments, the sensor platform is ideally positioned at the center of the landing pad, with its height aligned to the pad’s, ensuring that the drone remains within the event camera’s FoV at all times for continuous localization.
}

\section{Evaluation} \label{6}

Our evaluation of mmE-Loc is comprehensive and grounded entirely in real-world experimentation.  
We begin with experimental settings in §\ref{5.1}. 
Then, in §\ref{5.2}, we focus on key performance metrics: localization accuracy and latency. 
§\ref{5.3} delves into external factors impacting mmE-Loc, such as drone characteristics and environmental conditions. 
§\ref{5.4} assesses benefits of sensor fusion and the performance of various modules. 
Finally, in §\ref{5.5}, we evaluate system load, including latency, CPU usage, and memory consumption.


\vspace{-0.3cm}
\subsection{Experimental methodology} \label{5.1}
\begin{figure*}[t]
    \centering
        \includegraphics[width=2.05\columnwidth]{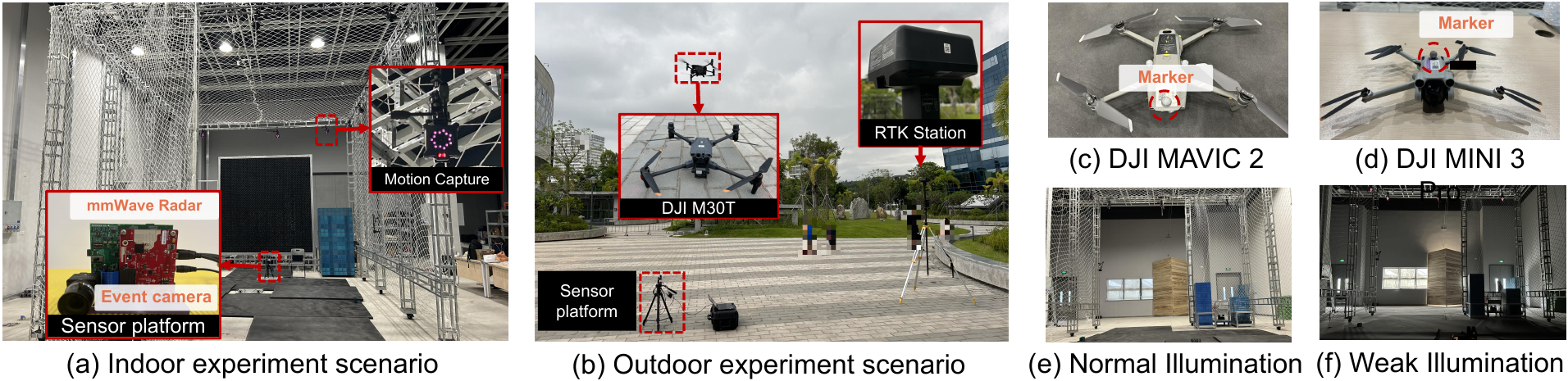}
        \vspace{-0.65cm}
    \caption{Experimental setup and scenarios of mmE-Loc. \textnormal{(a) A laboratory scenario with motion capture system for ground-truth collection. (b) An outdoor scenario with RTK system for ground-truth collection. (c)-(d) Different drone with different size (DJI MAVIC 2 and DJI MINI 3 Pro). (e)-(f) Different illumination situation of the laboratory (normal and weak).}}
    \label{setup}
    \vspace{-0.4cm}
\end{figure*}

\RR{
\textbf{Experiment setting.} 
We use a deployment setup and drone equipment that closely mimic real-world applications. 
\fig \ref{setup} shows the experimental scenarios in both an \textit{indoor} laboratory and an \textit{outdoor} flight test site. 
The setup in \fig \ref{setup}a includes the event camera and mmWave radar mounted together \textit{at the ground of the experimental area}.
We evaluate mmE-Loc on various target drones:
$(i)$\textit{ DJI Mini 3 Pro}: 0.25$m$ $\times$ 0.36$m$ $\times$ 0.07$m$ with unfolded propellers.
$(ii)$ \textit{DJI MAVIC 2}: 0.32$m$ $\times$ 0.24$m$ $\times$ 0.08$m$ with unfolded propellers.
$(iii)$ \textit{DJI M30T}: 0.49$m$ $\times$ 0.61$m$ $\times$ 0.22$m$ with unfolded propellers.
By default, in the indoor experiments, the drone flew above the sensor platform at an altitude of \(3\text{--}4~\text{m}\), following a zigzag trajectory that covered an area of \(2~\text{m} \times 3~\text{m}\).  
This setup ensured that the system could continuously localize the drone during its final landing phase, while the motion capture system simultaneously recorded ground-truth trajectories.  
In the outdoor experiments, the drone flew at an altitude of \(5\text{--}~6\text{m}\), covering the same \(2~\text{m} \times 3~\text{m}\) region with a similar zigzag pattern to maintain reliable localization until touchdown.  
The DJI Mini 3 Pro drone was used in the indoor and robustness experiments conducted under controlled conditions, whereas the DJI M30T drone was employed for the outdoor experiments.  
A DJI Mavic 2 drone was introduced to evaluate the system’s localization performance across different drone models.
We conducted extensive experiments over \textit{30} hours, collecting \textit{more than 400GB} of raw data.
}


\RR{
\textbf{Ground truth.}
In the indoor scenario, we use a NOKOV motion capture system equipped with fourteen cameras. 
The system covers an \textit{8~m~$\times$~8~m~$\times$~8~m} area and provides localization accuracy within \textit{1~mm}, enabling precise performance evaluation under controlled conditions.
In the outdoor scenario, we conduct experiments at an open test site with excellent GPS signal reception. 
A DJI D-RTK2 Real-Time Kinematic (RTK) base station is deployed to provide carrier-phase corrections to the Matrice~M30T drone, ensuring high-precision positioning \cite{DJI}. 
The RTK system achieves a localization error of \textit{1~cm (horizontal)} and \textit{2~cm (vertical)}.
These RTK results were used as the ground truth for the outdoor experiments.
}

\begin{figure*}
\setlength{\abovecaptionskip}{-0.25cm} 
\setlength{\belowcaptionskip}{-0.15cm}
\setlength{\subfigcapskip}{-1cm}
  \begin{minipage}[t]{0.67\columnwidth}
    \centering
    \includegraphics[width=1\columnwidth]{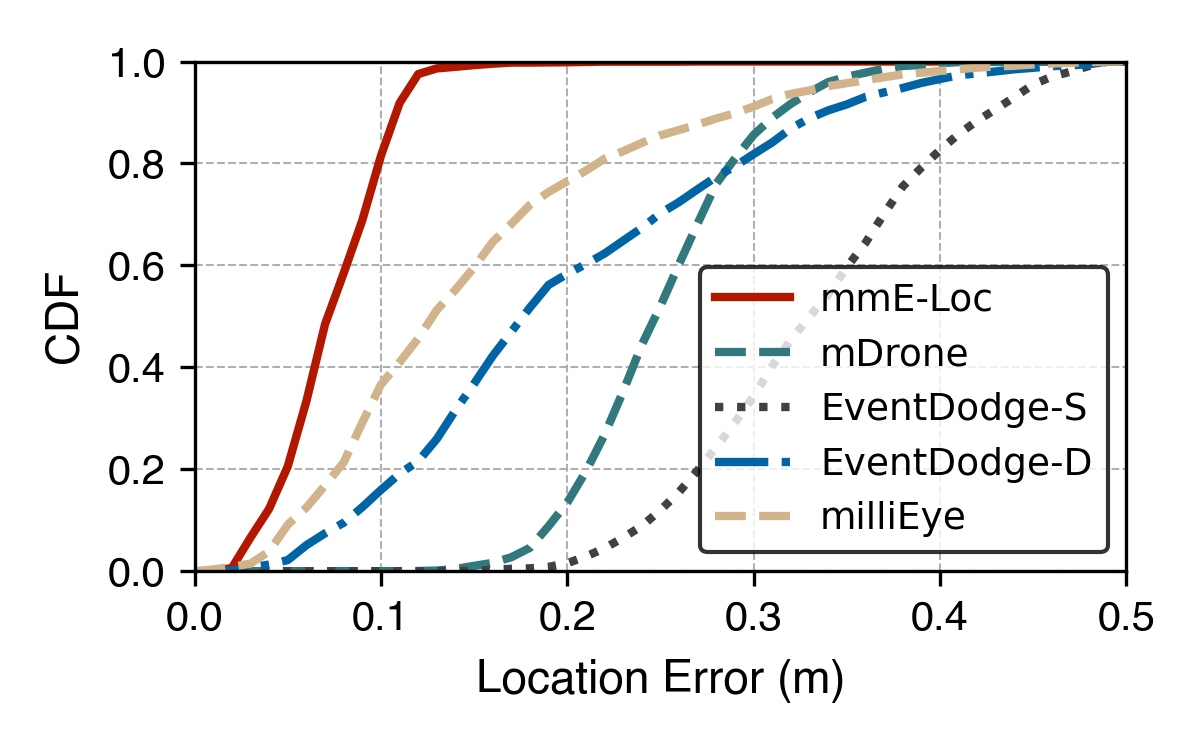}
    \caption{\RR{Indoor accuracy comparison}}
    \label{fig:indoor_cdf}
  \end{minipage}
  \begin{minipage}[t]{0.67\columnwidth}
    \centering
    \includegraphics[width=1\columnwidth]{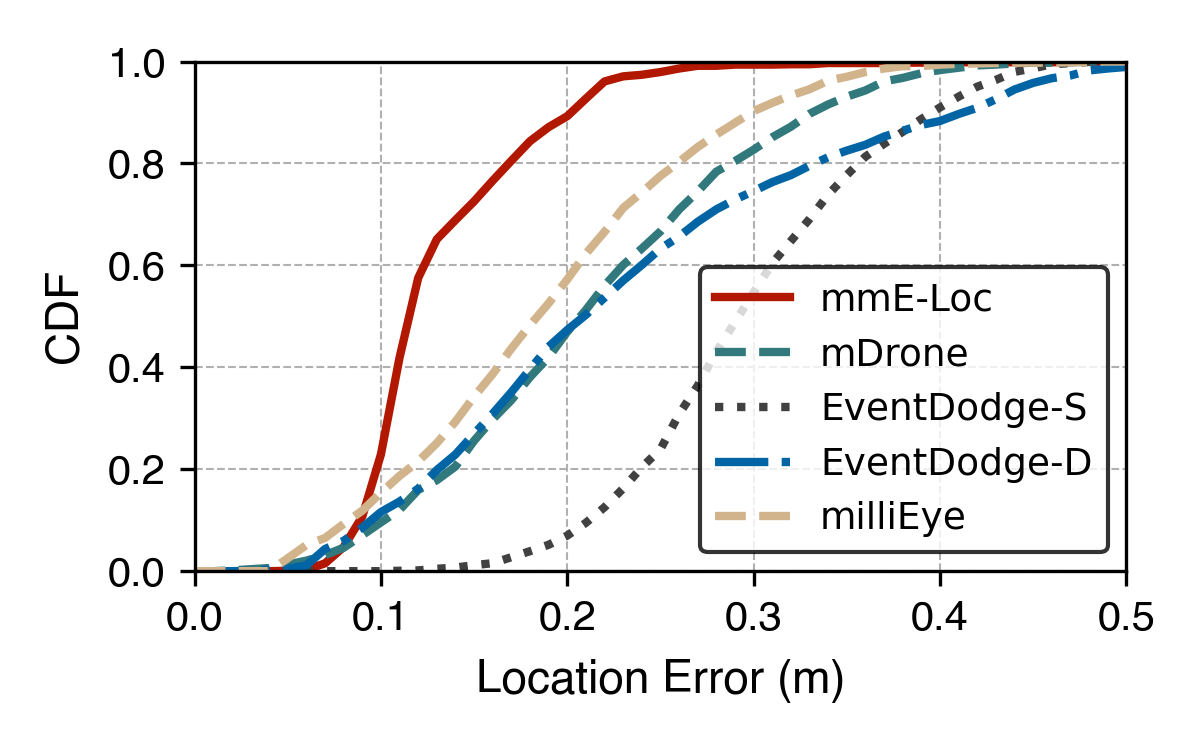}
    \caption{\RR{Outdoor accuracy comparison}}
    \label{fig:outdoor_cdf}
  \end{minipage}
  \begin{minipage}[t]{0.67\columnwidth}
    \centering
    \includegraphics[width=1\columnwidth]{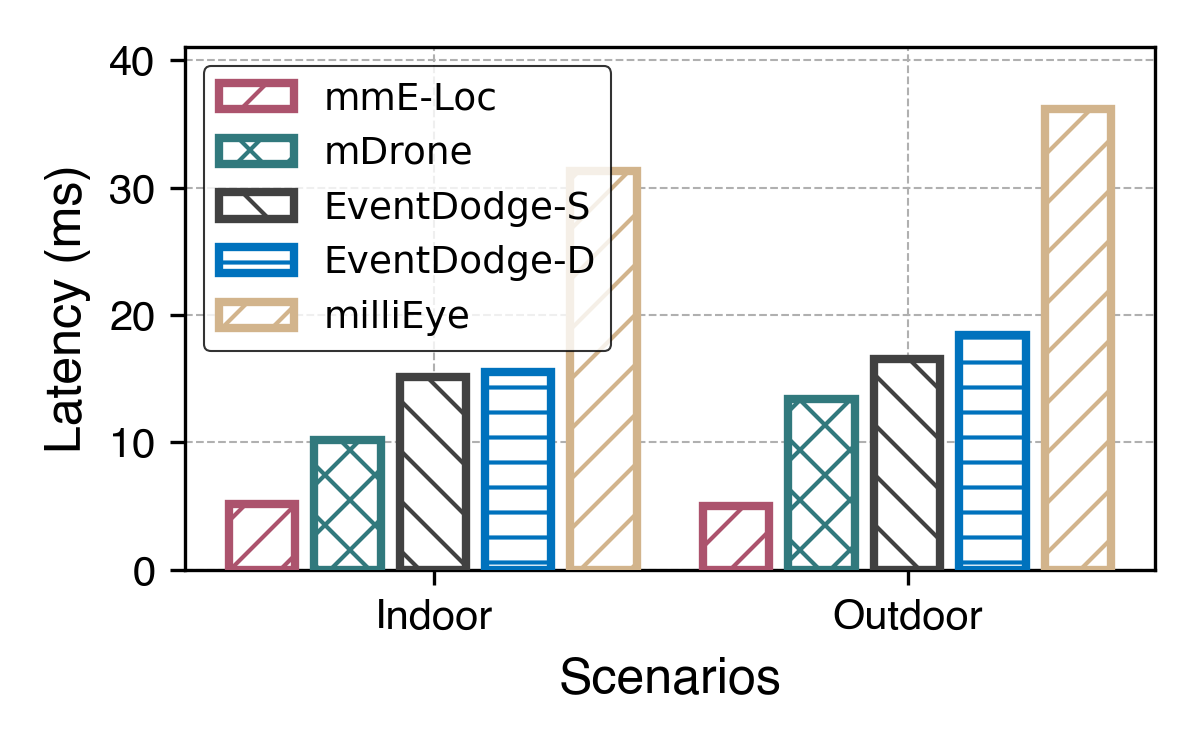}
    \caption{\RR{Latency comparison}}
    \label{fig:indoor_runtime}
  \end{minipage}
  \hfill
  \vspace{-0.3cm}
\end{figure*}

\RR{

\textbf{Comparative methods.}
We re-implement and adapt the following state-of-the-art (SOTA) algorithms to our experimental setup for fair comparison:
$(i)$ \textbf{mDrone} \cite{zhao20213d}: a SOTA point-cloud-based drone localization system using single-chip mmWave radar, originally designed for ground localization. We adjusted its parameters for our radar configuration and ground-based testing.
This method is chosen as a baseline to validate the effectiveness of using an event camera for 2D spatial calibration of the mmWave radar and its contribution to improving localization accuracy.
$(ii)$ \textbf{EventDodge-S} \cite{falanga2020dynamic}: a SOTA \textbf{S}ingle event camera-based onboard obstacle localization system applies to obstacles with known geometries.
Since the original code is not publicly available, we re-implemented its monocular 3D localization module and adapted it for \textit{ground-based} drone localization, excluding ego-motion compensation for noise reduction.
$(iii)$ \textbf{EventDodge-D} \cite{falanga2020dynamic}: a SOTA \textbf{D}ual event camera-based onboard obstacle localization system. 
The original method uses onboard stereo event cameras for obstacle localization, but its code is unavailable. 
We implement its stereo 3D location estimation module, excluding ego-motion compensation, for \textit{ground-based} drone localization.
These two methods are chosen as baselines to validate the effectiveness of introducing the mmWave radar for providing depth information to the event camera and to assess its contribution to the overall localization accuracy.
$(iv)$ \textbf{milliEye} \cite{shuai2021millieye}: a deep learning-based multimodal localization system that integrates monocular images and mmWave point clouds. 
We pre-trained the neural networks following the original design and incorporated a Kalman Filter to adapt the method for 3D \textit{ground-based} drone localization.
This method is selected as a baseline to demonstrate the significance of upgrading the frame camera to an event camera, showing that this upgrade leads to substantial improvements in both localization accuracy and latency.
}
\revise{
\textbf{Evaluation metrics.}
mmE-Loc provides continuous, low-latency location estimates, ensuring accurate and responsive drone localization.
Similar to related works, we evaluate location estimation error in meters and processing latency in milliseconds, facilitating direct comparison with existing methods in terms of both accuracy and efficiency.
}

\RR{
\textbf{Robustness experiments.}
During drone landing, localization may occur under varying conditions.
To assess mmE-Loc's adaptability, we conduct experiments across diverse scenarios, including different environments, drone models, lighting conditions (\fig \ref{setup}a-f), and background dynamics (\eg, moving objects like balls). 
We also evaluate the impact of distance, occlusions (\eg, partial obstruction of the drone in the event camera’s FoV), and velocity to demonstrate robustness.
It is worth noting that the robustness evaluation in our work primarily focuses on the operation scenarios of delivery drones or those performing drone-driven low-altitude economy activities. 
This means that our main objective is to guide cooperative drones to accurately and reliably land on the designated landing pad.
\textit{Owing to the considerable latency introduced by the frame camera’s exposure and readout processes (as illustrated in Fig.~\ref{intro}), milliEye is excluded from the robustness experiments. }
The resulting delay, typically tens of milliseconds, leads to poor temporal responsiveness under dynamic conditions. 
Since our objective is to enable accurate and low-latency drone localization, Baseline-IV does not align with the intended low-latency evaluation focus.
}

\vspace{-0.2cm}
\subsection{Overall performance} \label{5.2}

\textbf{Drone localization.} 
\fig \ref{fig:indoor_cdf} shows the localization performance of mmE-Loc in an indoor environment using a DJI Mini 3 Pro drone, compared to four baselines.
mmE-Loc achieves an average end-to-end localization error of 0.083$m$, outperforming the baselines of mDrone, EventDodge-S, EventDodge-D, and milliEye with errors of 0.261$m$, 0.345$m$, 0.209$m$, and 0.160$m$, making it well-suited for landing assistance.
\fig \ref{fig:outdoor_cdf}b presents mmE-Loc’s localization performance in an outdoor setting with a DJI M30T drone. Here, mmE-Loc achieves the lowest average error of 0.135$m$, outperforming the baselines by 39.2\%, 56.0\%, 43.8\%, and 31.8\%, respectively.
mDrone suffers from point cloud errors caused by phase center offset, resulting in deviations from the drone’s geometric center.
In indoor environments, these errors are exacerbated by specular reflections, diffraction, and multi-path effects that further degrade radar measurements.
EventDodge-S employs a single event camera with prior knowledge of drone geometry, while EventDodge-D uses a stereo setup with dual event cameras for depth estimation. Both approaches are vulnerable to environmental noise (e.g., birds), which can lead to inaccurate depth estimation.
milliEye relies on deep learning models, which may exhibit higher errors when deployed in unseen environments due to limited generalization.
In contrast, mmE-Loc significantly improves performance by leveraging the complementary strengths of radar and event cameras. It operates without the need for pre-training, enabling robust and broadly applicable localization.
A more detailed analysis is provided in Appendix F.

\textbf{End-to-end latency.}
We evaluate end-to-end latency, including the \textit{CCT} and \textit{GAJO} phases.
As shown in \fig \ref{fig:indoor_runtime}, mmE-Loc achieves 5.12$ms$ latency indoors, outperforming baselines of mDrone, EventDodge-S, EventDodge-D, and milliEye at 10.19$ms$, 15.12$ms$, 15.52$ms$, and 31.2$ms$, respectively. 
In outdoor scenarios with higher complexity, baseline latencies increase, and mmE-Loc maintains the lowest latency, outperforming them by 62.5\%, 69.6\%, 72.7\%, and 83.3\%. 
Baselines face increased latency due to environmental noise and more optimization parameters.
mmE-Loc establishes a tight coupling between the event camera and mmWave radar.
It incorporates \textit{CCT} module to exploit temporal consistency, drone's  periodic micro-motions and structure for drone detection. 
Then, \textit{GAJO} module, combined with an incremental optimization approach, enhances spatial complementarity while minimizing system latency.

\RR{
\textbf{Comparison with Lidar-based method.}
To further demonstrate the advantages of mmE-Loc, we conducted indoor experiments using the same setup as mmE-Loc and a DJI Mini 3 Pro drone (measuring 0.25 m × 0.36 m × 0.07 m with unfolded propellers). 
In this experiment, we compared the localization accuracy and latency of mmE-Loc against a LiDAR-based method \cite{vrba2024onboard}. 
\cite{vrba2024onboard} leverages onboard LiDAR to detect and localize flying objects. Specifically, its detector employs 3D voxel mapping for object detection, achieving high localization accuracy and robustness to environmental variations. 
Furthermore, the method integrates a clustering-based tracker to suppress false positives and to estimate and predict the target’s state.
For a fair comparison with mmE-Loc, we configured \cite{vrba2024onboard}’s observer as static, detecting and localizing a flying drone to simulate a drone landing scenario.
The LiDAR used was a RoboSense Helios-16-line LiDAR, priced at approximately \$3,000, with a scanning rate of 10 Hz. Both methods were executed on the same computing platform, a PC equipped with an Intel i7 CPU.
When using only the detector, the LiDAR-based method achieved an average localization error of 0.071 $m$ and an average latency of 658.82 $ms$. 
After integrating the tracker, the average localization error increased to 0.165 $m$, while the latency decreased to 29.74 $ms$. 
This trade-off occurs because the tracker predicts the target’s position, thereby reducing latency but inevitably introducing additional estimation errors. 
The residual localization error also arises from the LiDAR’s limited field of view, as each scan only captures the target from one side, leading to uneven point cloud sampling and a positional bias within the range of $[0, a]$, where $a$ denotes the radius of the target’s bounding sphere.
Such behavior aligns well with the description in \cite{vrba2024onboard}.
As shown by the results, although incorporating LiDAR enables a localization accuracy comparable to that of mmE-Loc, its latency remains more than 5x higher, even with tracker assistance. 
In realistic landing scenarios, achieving both rapid and reliable localization updates is critical.
Hence, a hybrid deployment that combines the strengths of the LiDAR-based method and mmE-Loc could be a promising solution. 
The LiDAR-based approach can offer precise position estimation, while mmE-Loc can complement it by providing low-latency updates between LiDAR scans, jointly supporting efficient drone landing guidance.
}



\begin{figure*}
\setlength{\abovecaptionskip}{-0.15cm} 
\setlength{\belowcaptionskip}{-0.2cm}
\setlength{\subfigcapskip}{-0.25cm}
  \begin{minipage}[t]{0.67\columnwidth}
    \centering
    \includegraphics[width=1\columnwidth]{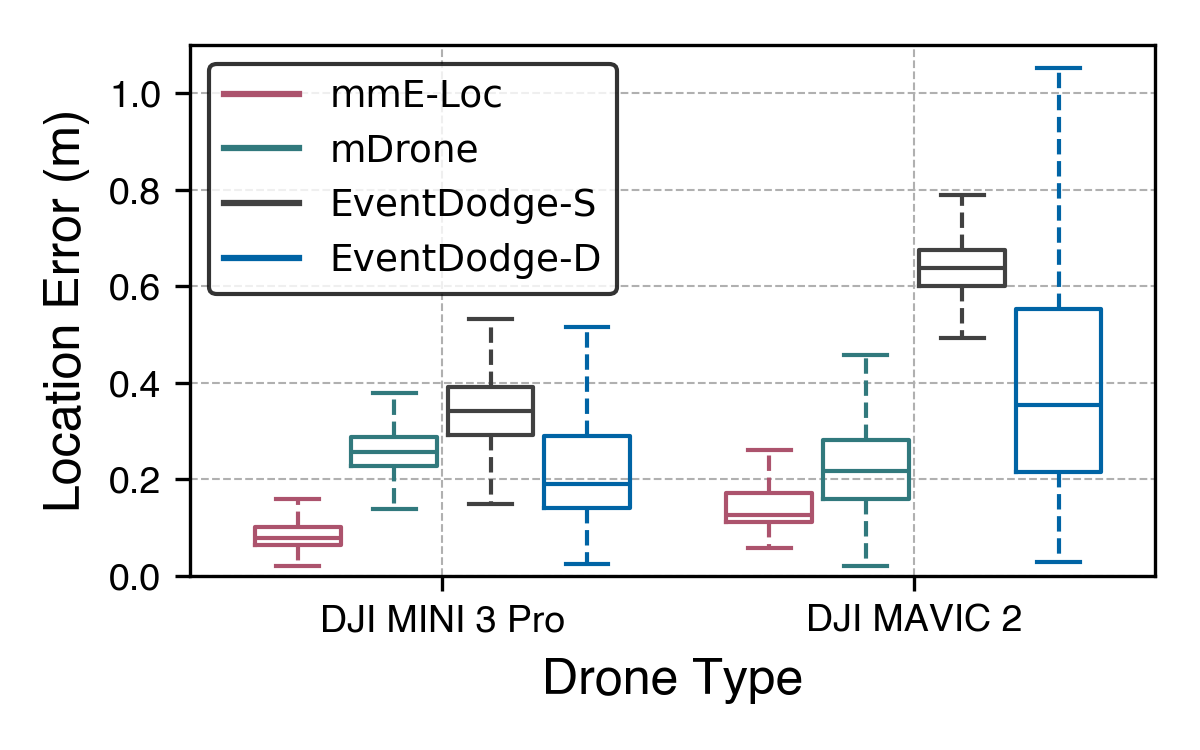}
    \caption{\RR{Impact of drone type}}
    \label{fig:drone_type}
  \end{minipage}
  \begin{minipage}[t]{0.67\columnwidth}
    \centering
    \includegraphics[width=1\columnwidth]{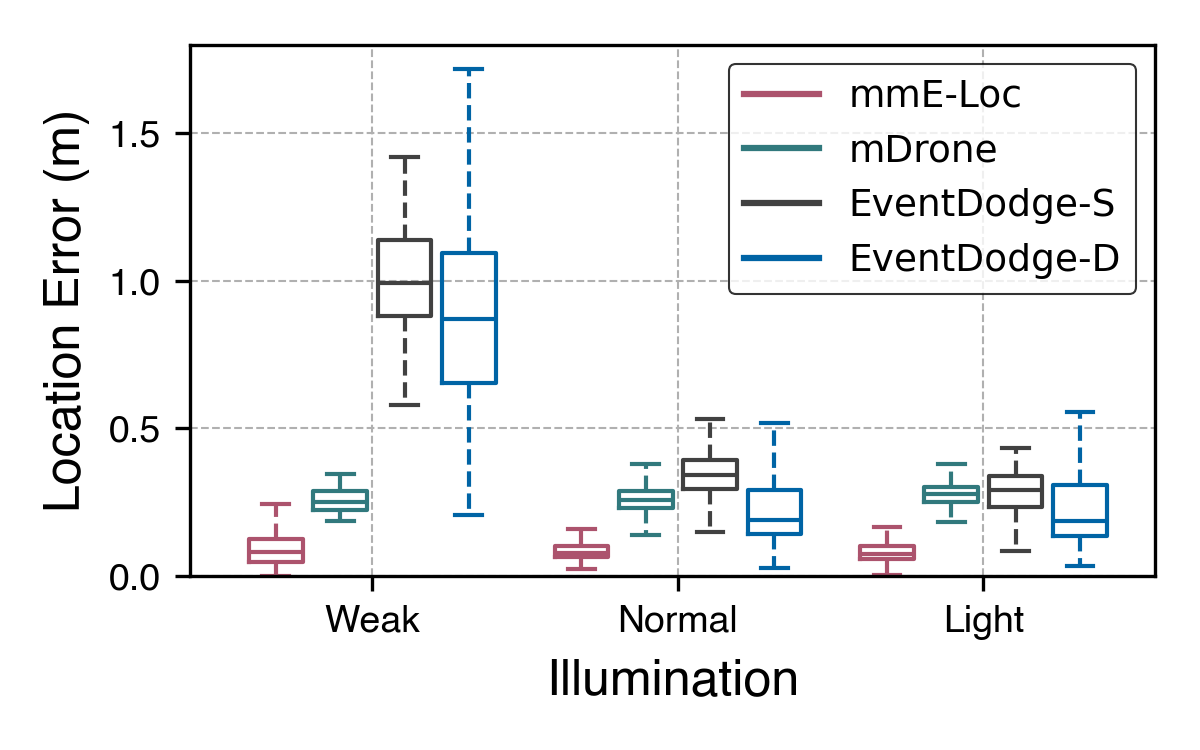}
    \caption{\RR{Impact of env. \& illu.}}
    \label{fig:illumination}
  \end{minipage}
  \begin{minipage}[t]{0.67\columnwidth}
    \centering
    \includegraphics[width=1\columnwidth]{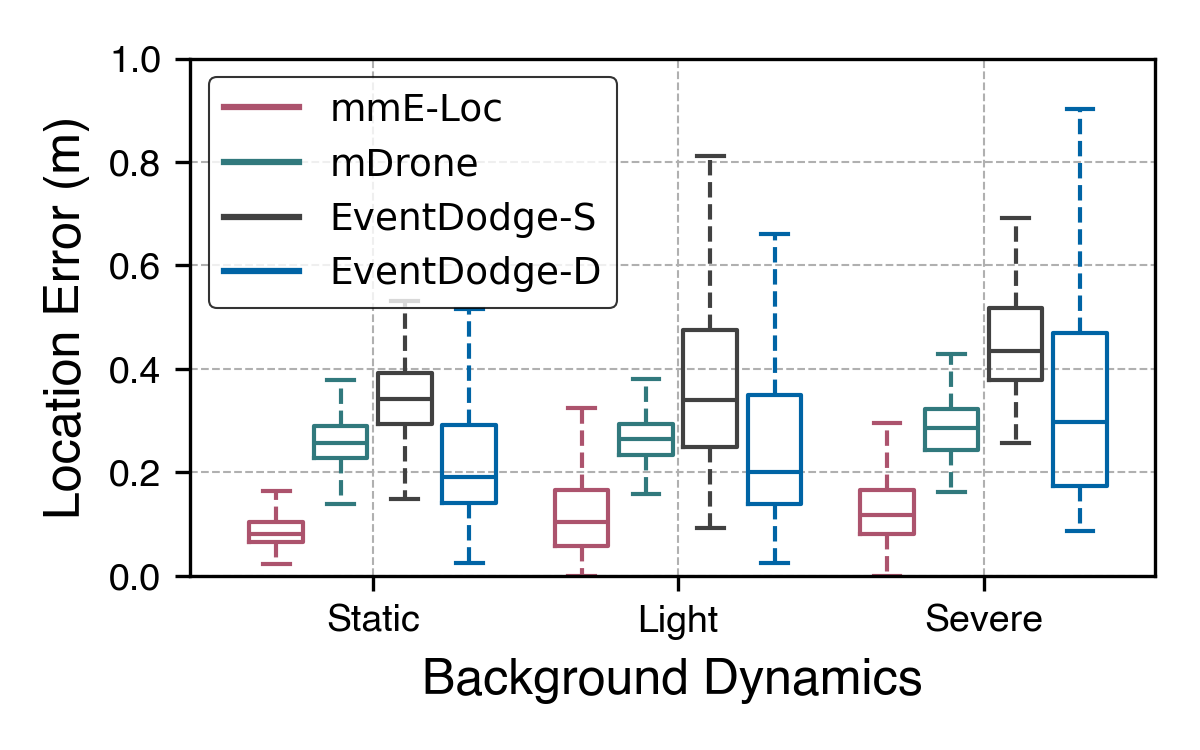}
    \caption{\RR{Impact of background}}
    \label{fig:background}
  \end{minipage}
  \hfill
  \vspace{-0.3cm}
\end{figure*}

\begin{figure*}
\setlength{\abovecaptionskip}{-0.15cm} 
\setlength{\belowcaptionskip}{-0.2cm}
\setlength{\subfigcapskip}{-0.25cm}
  \begin{minipage}[t]{0.67\columnwidth}
    \centering
    \includegraphics[width=1\columnwidth]{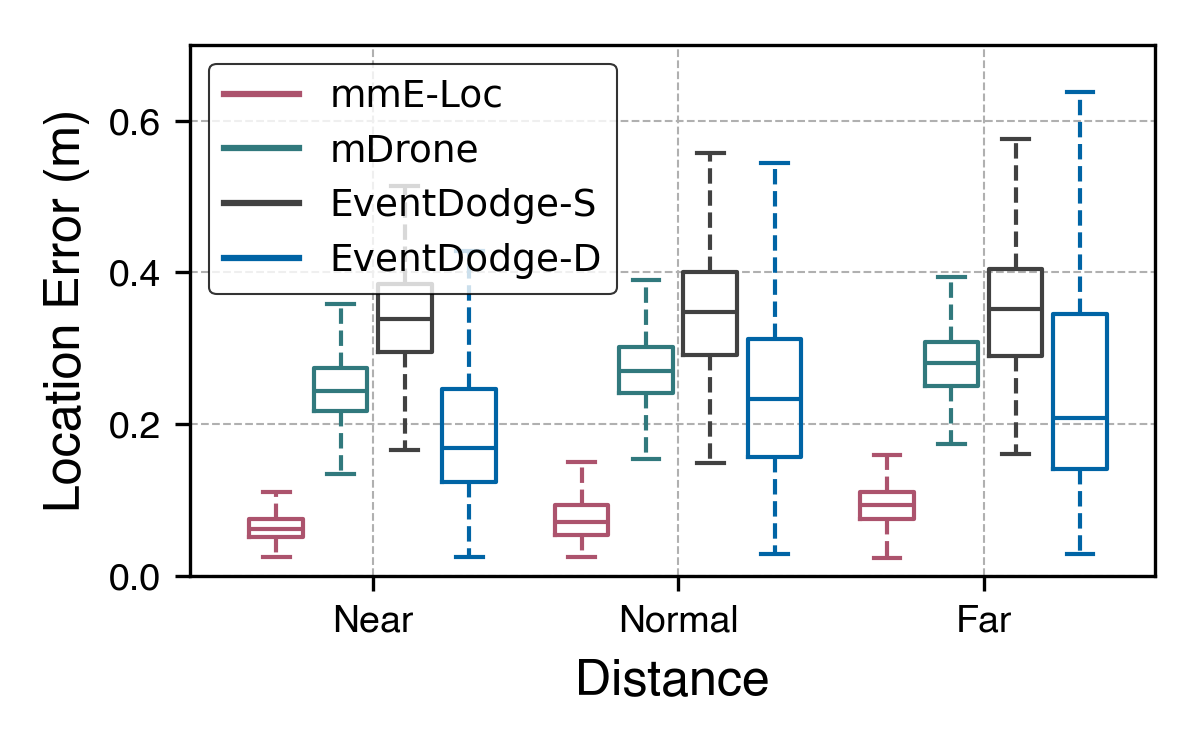}
    \caption{\RR{Impact of distance}}
    \label{fig:distance}
  \end{minipage}
  \begin{minipage}[t]{0.67\columnwidth}
    \centering
    \includegraphics[width=1\columnwidth]{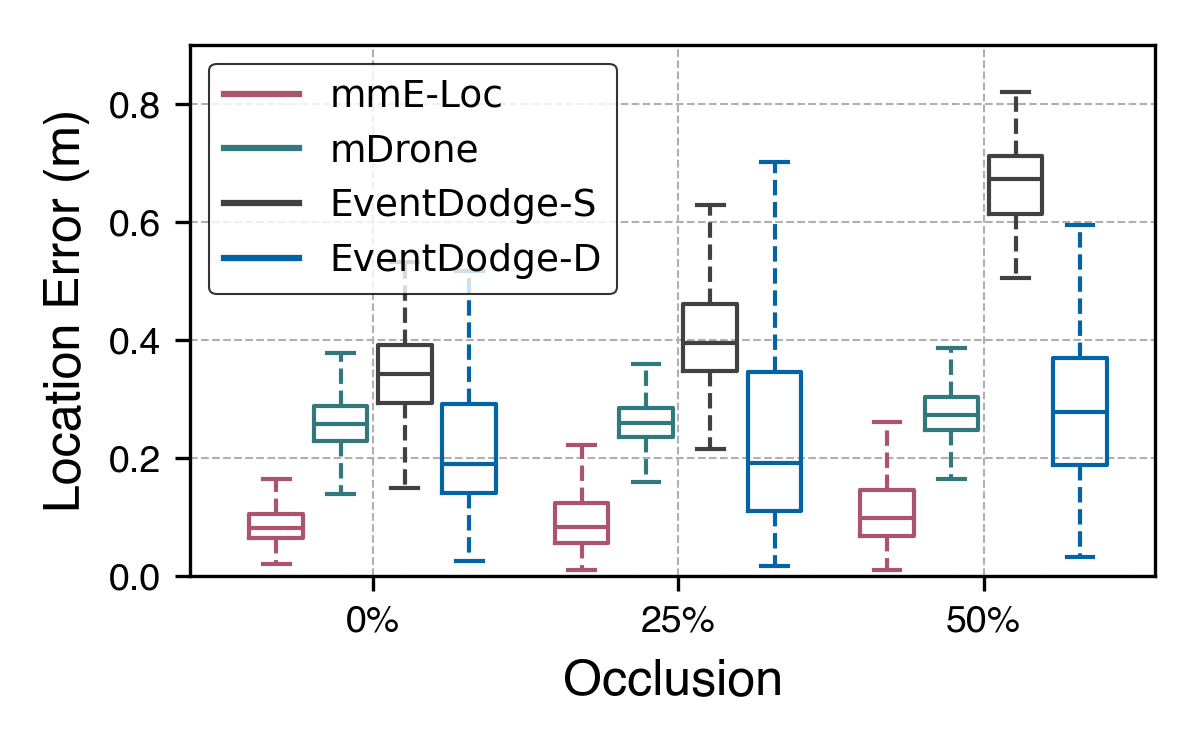}
    \caption{\RR{Impact of occlusion}}
    \label{fig:occlusion}
  \end{minipage}
  \begin{minipage}[t]{0.67\columnwidth}
    \centering
    \includegraphics[width=1\columnwidth]{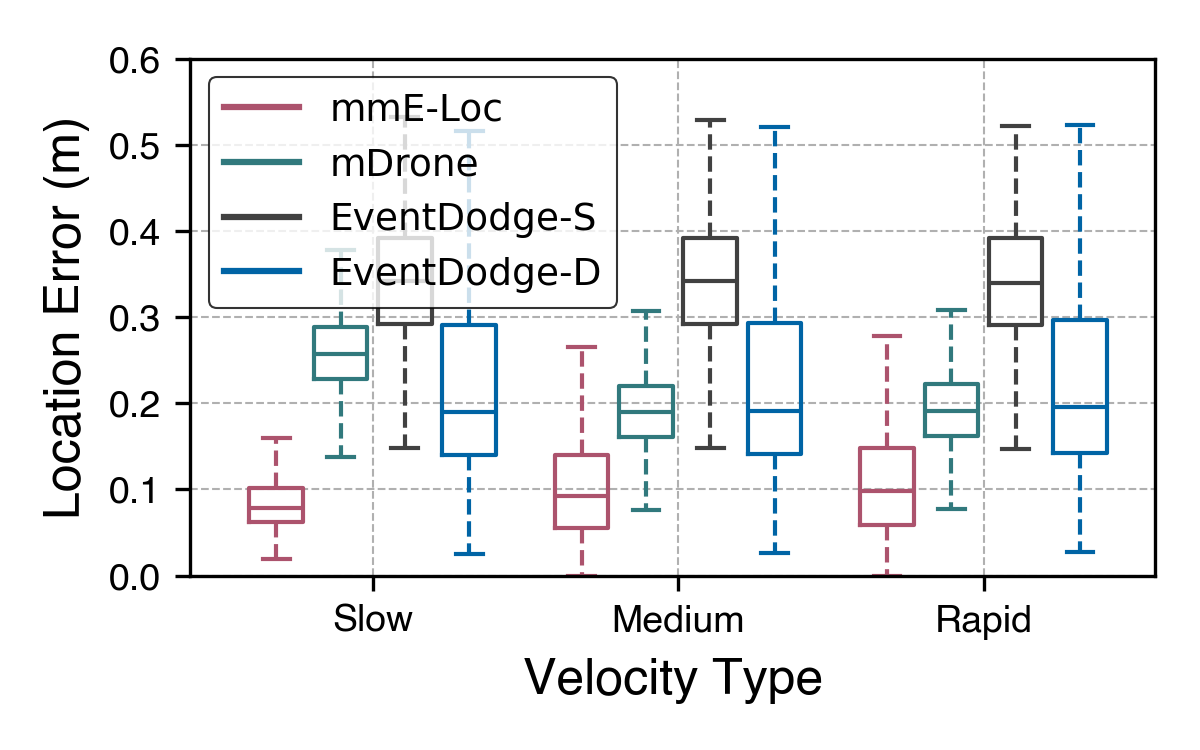}
    \caption{\RR{Impact of velocity}}
    \label{fig:velocity}
  \end{minipage}
  \hfill
  \vspace{-0.5cm}
\end{figure*}

\vspace{-0.3cm}
\subsection{System robustness evaluation} \label{5.3}
To demonstrate the versatility of mmE-Loc, we conduct experiments under various conditions.

\textbf{Impact of drone type.} 
\TMCrevise{
We evaluate the impact of different drone types under controlled indoor conditions, using a DJI Mini 3 Pro (\fig \ref{setup}c) and a DJI Mavic 2 (\fig \ref{setup}d), the results are presented in \fig \ref{fig:drone_type}. 
The average error results demonstrate that mmE-Loc outperforms all baselines when using DJI Mini 3 Pro.
For DJI Mavic 2, EventDodge-S and EventDodge-D exhibit larger localization errors, primarily due to changes in drone's geometry that impact depth estimation.
mmE-Loc's error of 0.135$m$ remains within an acceptable range, which outperforms all baselines with 0.222$m$, 0.639$m$, and 0.403$m$, demonstrating its effectiveness across drone types.
}

\RR{
\textbf{Impact of environment \& illumination.}
The performance of mmE-Loc in different environments with varying lighting conditions (\fig \ref{setup}e and \fig \ref{setup}f) is shown in \fig \ref{fig:illumination}. 
The experiments were conducted in an indoor environment, where environmental variables could be precisely controlled. 
The experimental settings are defined as follows: Weak refers to an indoor scene with an illumination level of approximately 50 lux, Normal corresponds to about 400 lux, and Light indicates a bright indoor condition of around 1000 lux. 
In outdoor scenarios, the ambient illumination level during drone delivery tasks generally exceeds 300 lux, indicating that the defined lighting condition effectively encompasses the illumination range commonly experienced in real-world drone delivery environments.
Although event cameras offer a high dynamic range, the events generated by drones are still influenced by illumination conditions.
As lighting diminishes, EventDodge-S and EventDodge-D suffer from increased localization errors due to degraded depth estimation performance.
In contrast, mDrone shows little variation in performance, as mmWave signals are inherently robust to illumination changes.
In comparison, mmE-Loc sustains a relatively low average error even under low illumination, recording an average error of 0.103$m$ and a maximum error of 0.27$m$.
This can be attributed to two main reasons:
$(i)$ The depth estimation in mmE-Loc primarily relies on the mmWave radar, which is unaffected by lighting conditions. In other words, mmE-Loc assigns greater weight to mmWave data when computing depth.
$(ii)$ The event camera mainly perceives the drone in the 2D spatial domain. 
Under weak illumination, its bounding box center may deviate from the drone’s true geometric center, leading to an increase in localization error.
Nevertheless, by detecting the periodic micro-motion of the drone, mmE-Loc can still produce accurate bounding boxes, enabling effective spatial calibration for the mmWave radar.
As a result, even in low-light conditions, the maximum error remains within a reasonable range and still outperforms localization using mmWave radar alone.
mmE-Loc's consistent performance across different lighting conditions, without requiring pre-training or prior knowledge, makes it a versatile solution.
}



\RR{
\textbf{Impact of background motion.}
We further evaluate mmE-Loc under dynamic background motion, as illustrated in \fig\ref{fig:background}.
The level of background dynamics is controlled by introducing moving objects of different sizes into the scene (e.g., a tennis ball with a diameter of approximately 7 cm and a football with a diameter of approximately 20 cm).
Specifically, in the Static condition, no other moving objects are present besides the drone; in the Light condition, a tennis ball is repeatedly tossed above the drone; and in the Severe condition, both a tennis ball and a football are continuously thrown across the drone’s flight area.
In real-world delivery scenarios, the disturbances that affect drone landings are predominantly caused by pedestrians, most commonly through object-throwing or transient movements. To emulate such realistic interference, our experimental setup is designed to cover representative forms of background motion frequently encountered during drone delivery landings.
As shown in the results, all baselines exhibit increased localization errors as background motion intensifies, primarily due to radar and vision misidentifications.
Under the Light and Severe conditions, EventDodge-S and EventDodge-D are more severely affected than mDrone, since overlapping trajectories between the balls and the drone distort the event camera’s bounding box generation and degrade depth estimation accuracy.
In contrast, the mmWave radar in mmE-Loc remains more robust, as the drone’s propeller rotation induces micro-motions at a much higher frequency than the movement of the balls, making the radar response more sensitive to the drone.
Despite the strong interference, mmE-Loc maintains an average localization error of only 0.129 m in the most challenging Severe scenario.
This robustness arises from its consistency-instructed measurement filter, which exploits the drone’s unique periodic micro-motion and structural priors to distinguish it from other moving objects.
If the background contains additional fast-moving objects (e.g., fans), mmE-Loc further analyzes the structural characteristics of the motion (whether other objects in the vicinity exhibit similar periodic micro-motions, and these objects possess symmetry) to determine whether the observed motion originates from the drone, thereby ensuring reliable tracking and localization.
}

\RR{
\textbf{Impact of distance.}
We investigate the impact of drone-to-platform distance indoors with a DJI Mini 3 Pro (0.25$m$ $\times$ 0.36$m$ $\times$ 0.07$m$). 
Since the application scenario of mmE-Loc focuses on landing guidance when a delivery drone approaches the landing pad, our experiments are conducted under three distance settings: Near ($<$3 m), Normal (3 m – 6 m), and Far ($>$6 m). These settings correspond to typical approach phases of a drone during descent and enable a comprehensive evaluation of mmE-Loc’s localization performance under different proximity conditions.
\fig \ref{fig:distance} shows the results.
As distance increases, all methods show higher errors. 
mmE-Loc achieves an average error of 0.102$m$ for far distances, outperforming other baselines. 
}
mmE-Loc combines mmWave radar for depth sensing with an event camera for high-resolution 2D imaging, effectively addressing the low spatial resolution of mmWave radar and the scale ambiguity of event cameras.
\TMCrevise{Recognizing the critical need for precise landings, mmE-Loc serves as a complementary system to existing solutions.
When integrated with RTK and visual markers, it enhances localization reliability and accuracy in real-world applications.}


\begin{figure*}
\setlength{\abovecaptionskip}{-0.2cm} 
\setlength{\belowcaptionskip}{-0.15cm}
\setlength{\subfigcapskip}{-1cm}
  \begin{minipage}[t]{0.5\columnwidth}
    \centering
    \includegraphics[width=1\columnwidth]{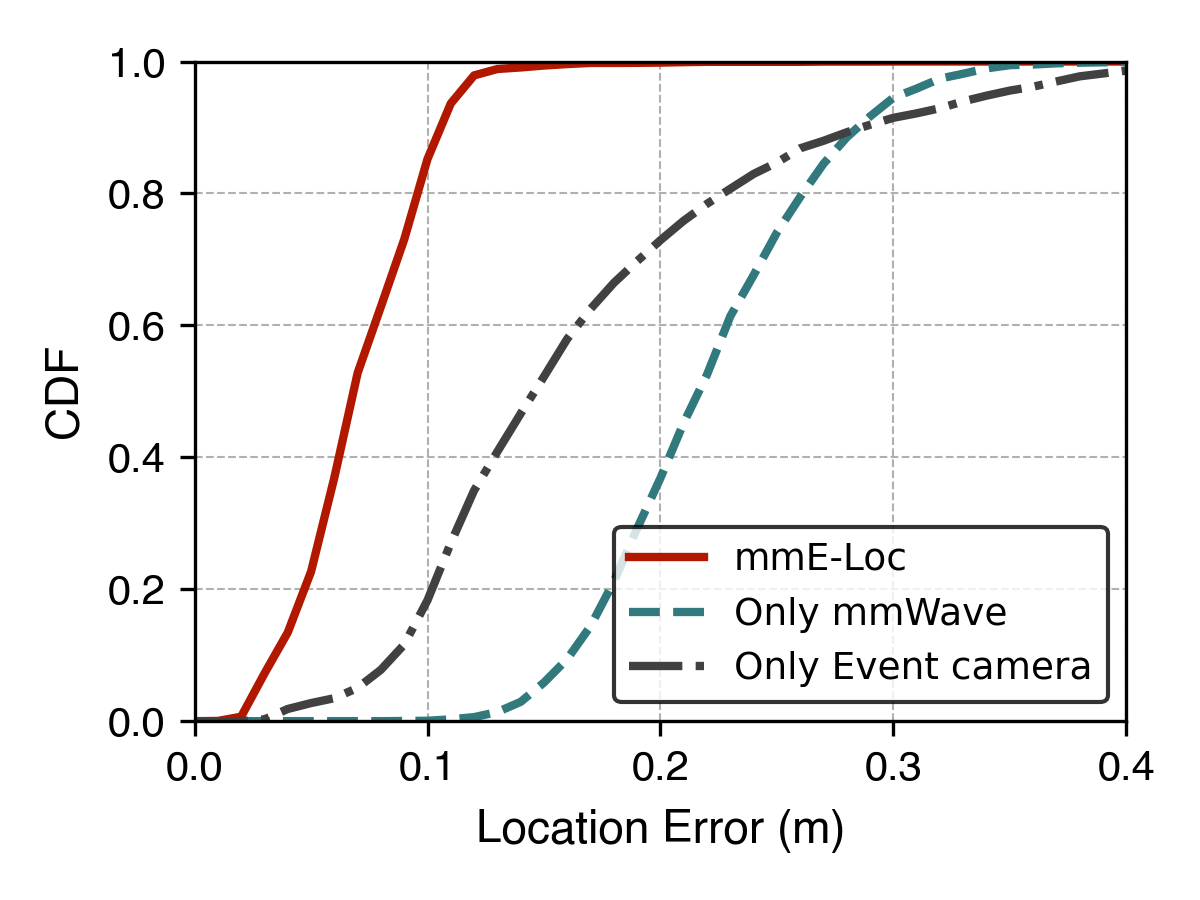}
    \caption{Effectiveness of sensor fusion}
    \label{fig:fusion}
  \end{minipage}
  \begin{minipage}[t]{0.5\columnwidth}
    \centering
    \includegraphics[width=1\columnwidth]{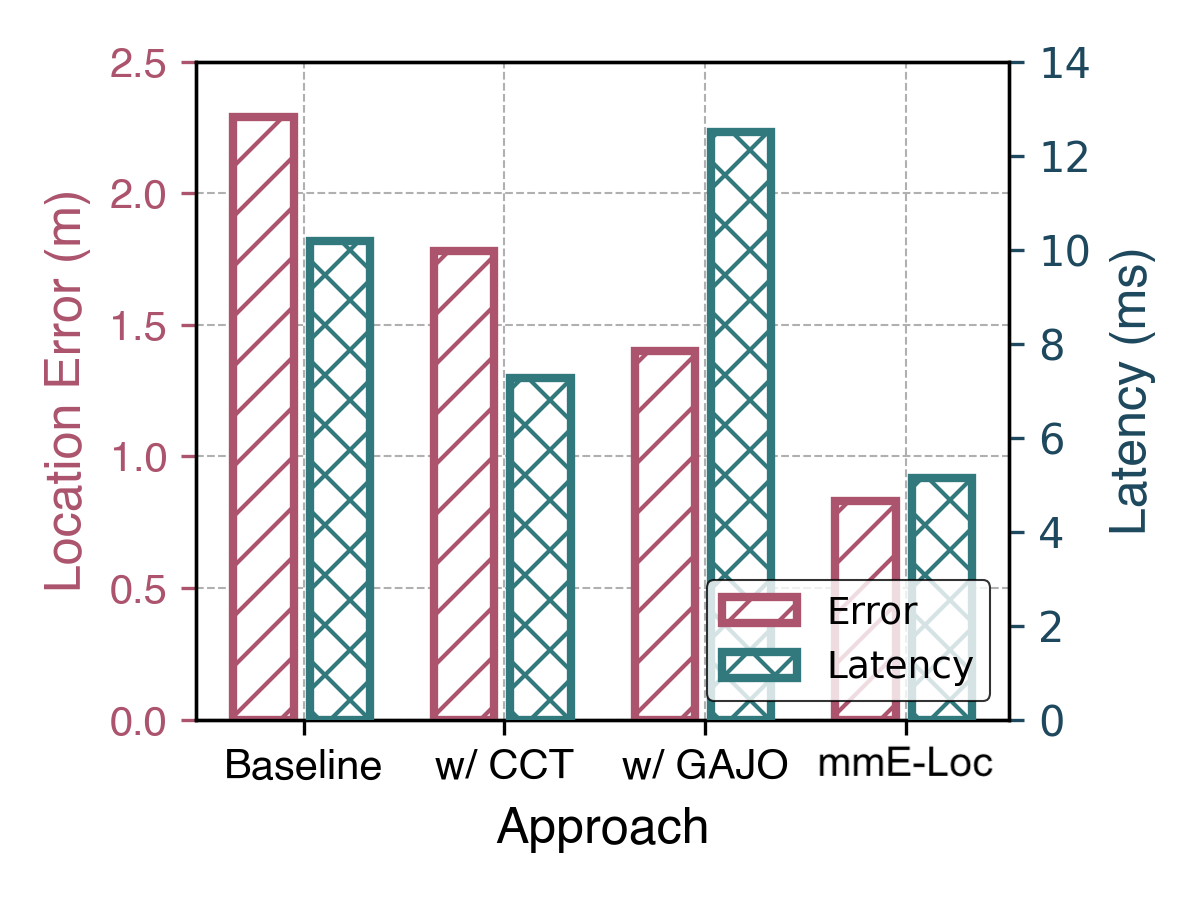}
    \caption{Impact of different modules}
    \label{fig:module}
  \end{minipage}
  \begin{minipage}[t]{0.5\columnwidth}
    \centering
    \includegraphics[width=1\columnwidth]{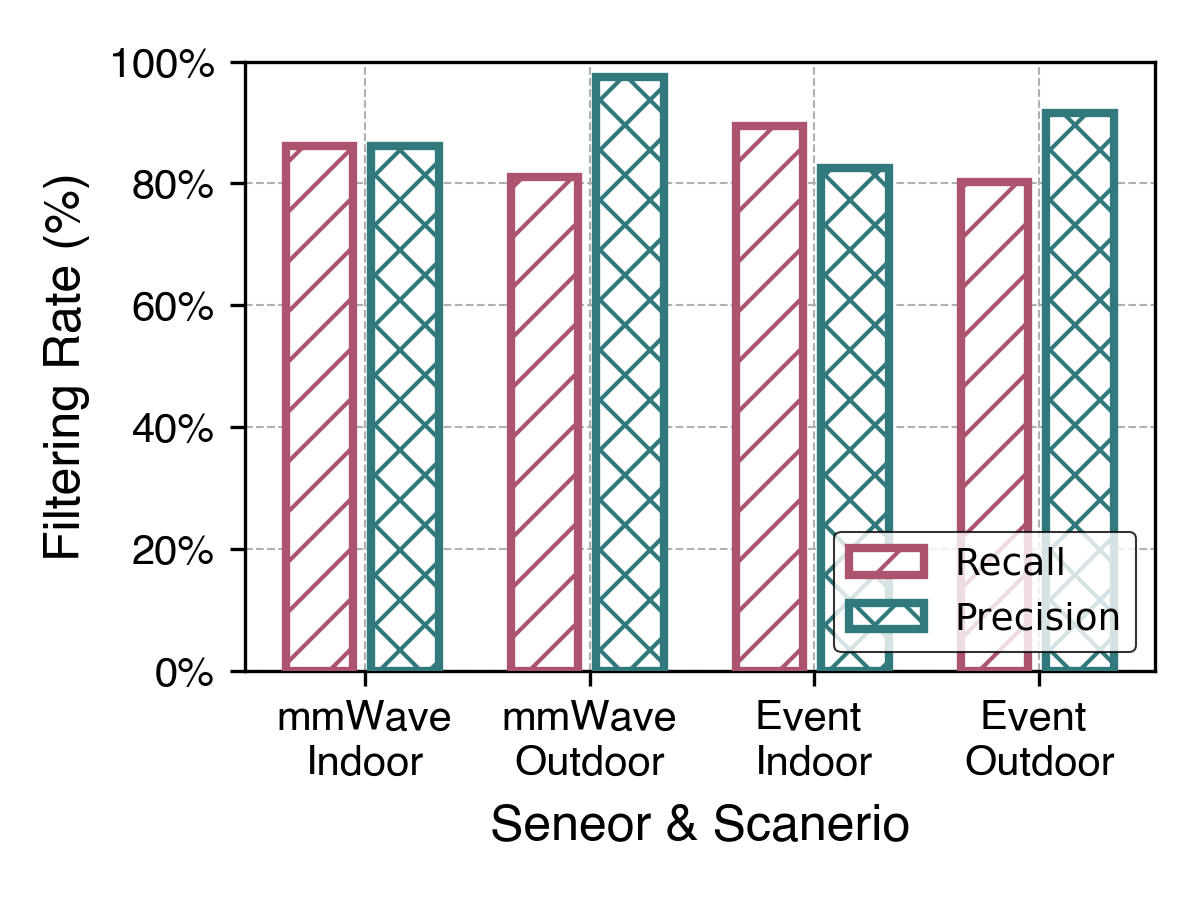}
    \caption{\RR{Performance of \textit{CCT}}}
    \label{fig:CCT}
  \end{minipage}
  \begin{minipage}[t]{0.5\columnwidth}
    \centering
    \includegraphics[width=1\columnwidth]{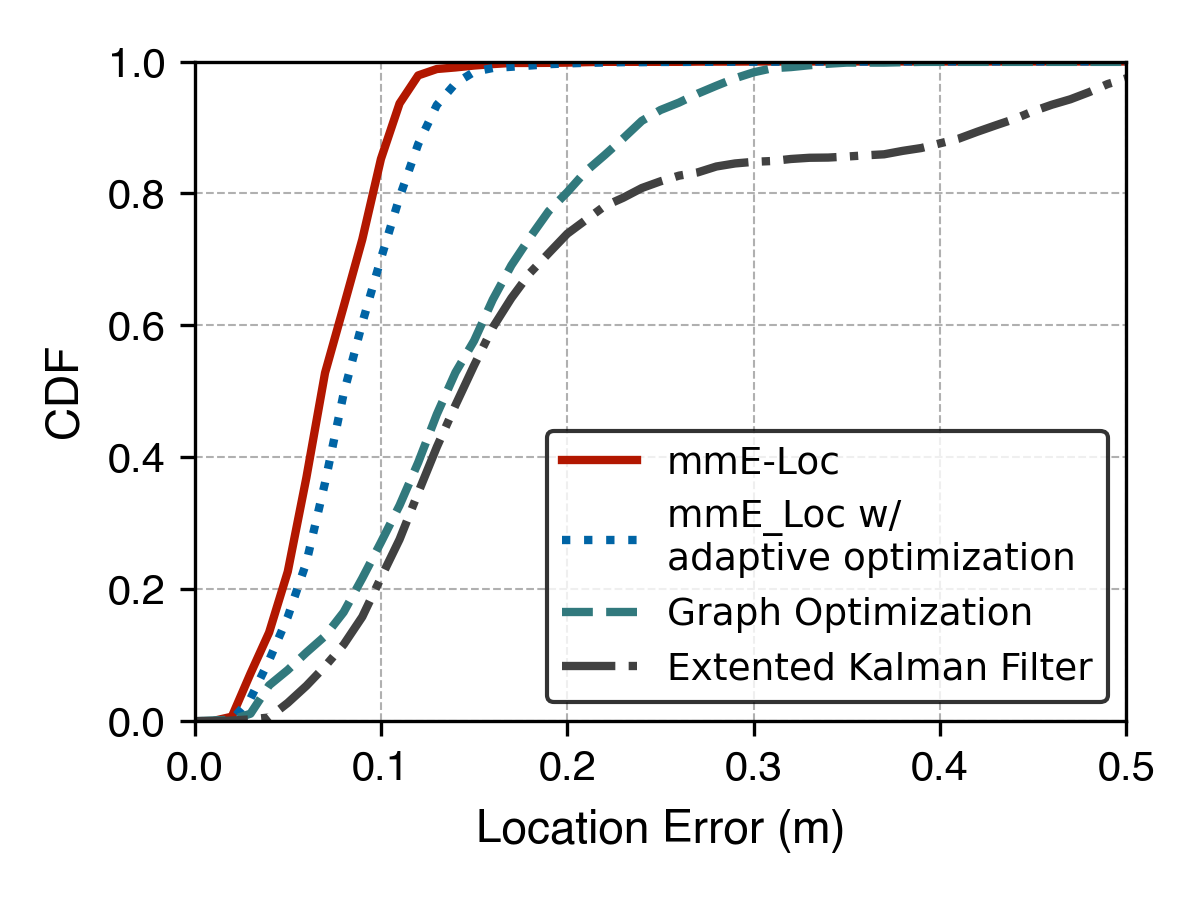}
    \caption{\RR{Performance of \textit{GAJO}}}
    \label{fig:GAJO}
  \end{minipage}
  \hfill
  \vspace{-0.4cm}
\end{figure*}

\begin{figure*}
\setlength{\abovecaptionskip}{-0.2cm} 
\setlength{\belowcaptionskip}{-0.15cm}
\setlength{\subfigcapskip}{-1cm}
  \begin{minipage}[t]{0.67\columnwidth}
    \centering
    \includegraphics[width=1\columnwidth]{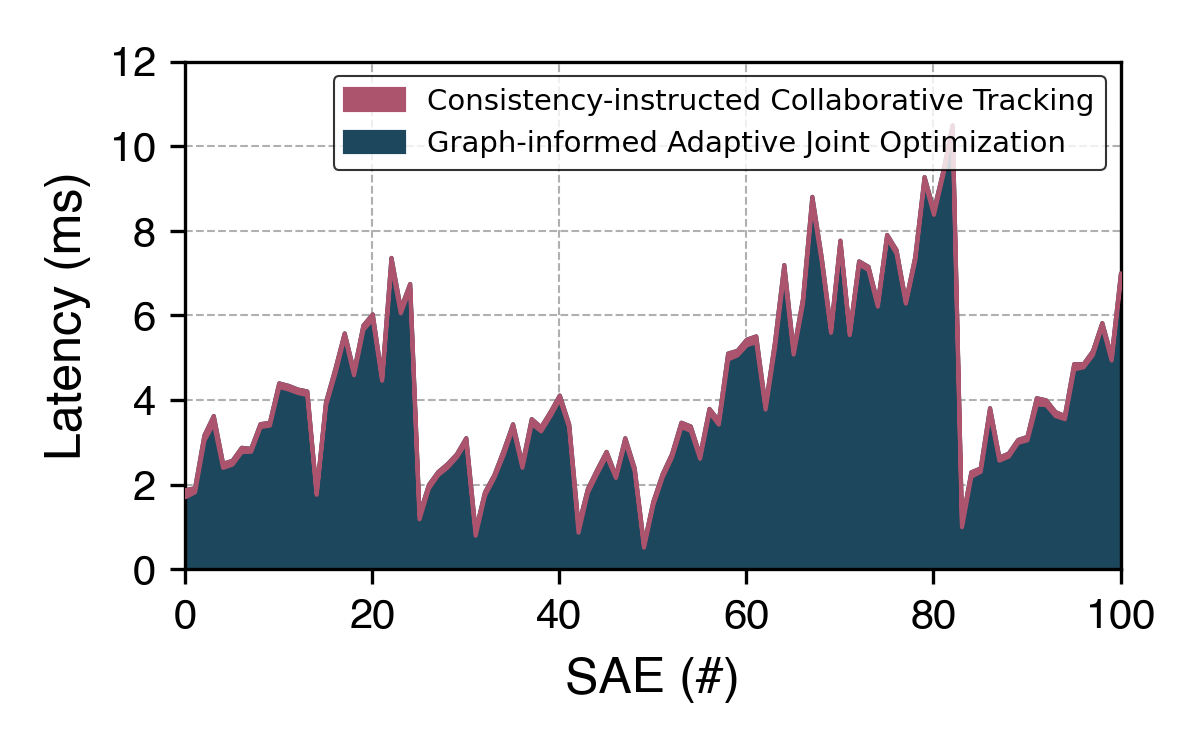}
    \caption{\TMCrevise{System latency}}
    \label{fig:latency}
  \end{minipage}
  \begin{minipage}[t]{0.67\columnwidth}
    \centering
    \includegraphics[width=1\columnwidth]{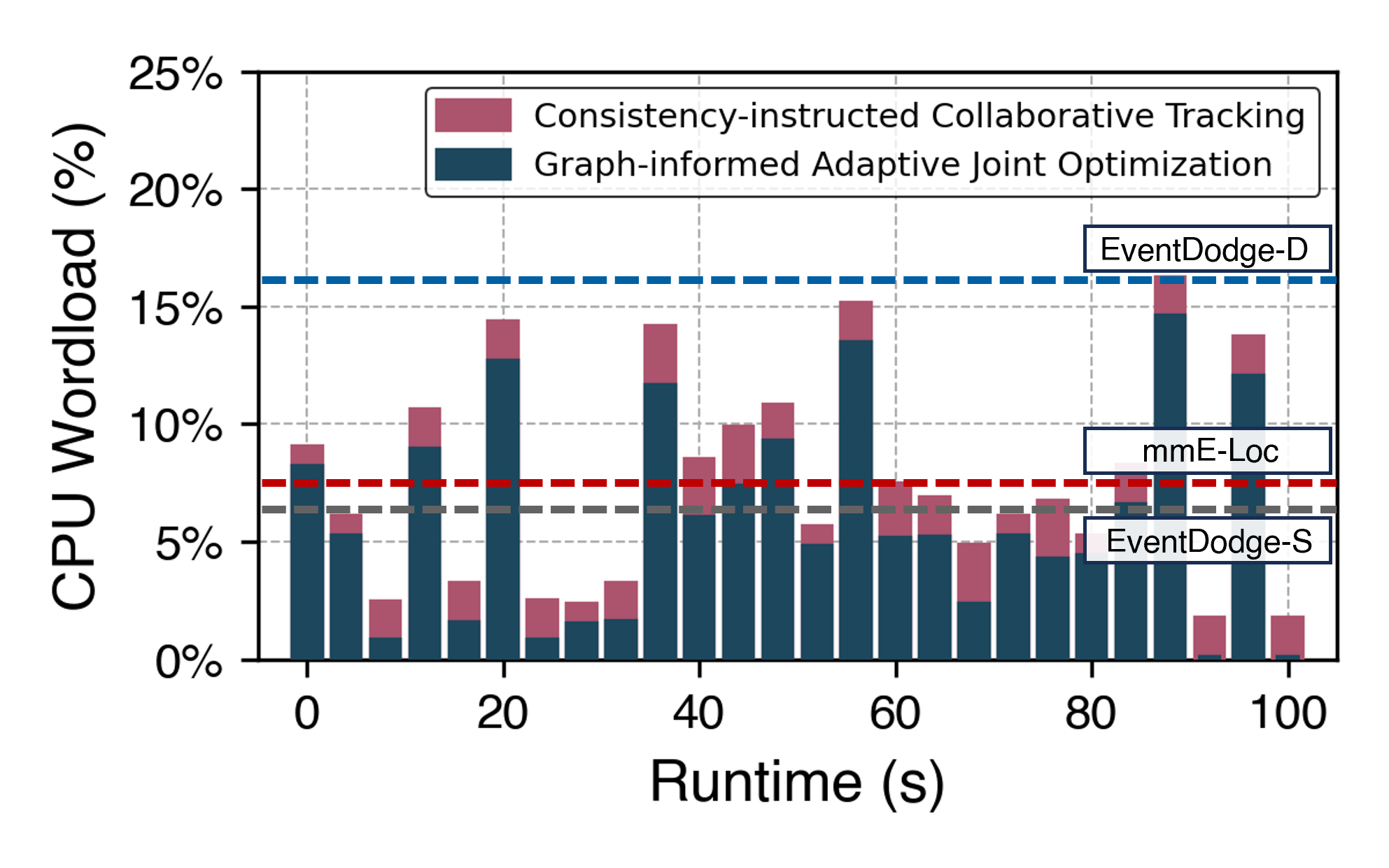}
    \caption{\RR{CPU workload}}
    \label{fig:cpu}
  \end{minipage}
  \begin{minipage}[t]{0.67\columnwidth}
    \centering
    \includegraphics[width=1\columnwidth]{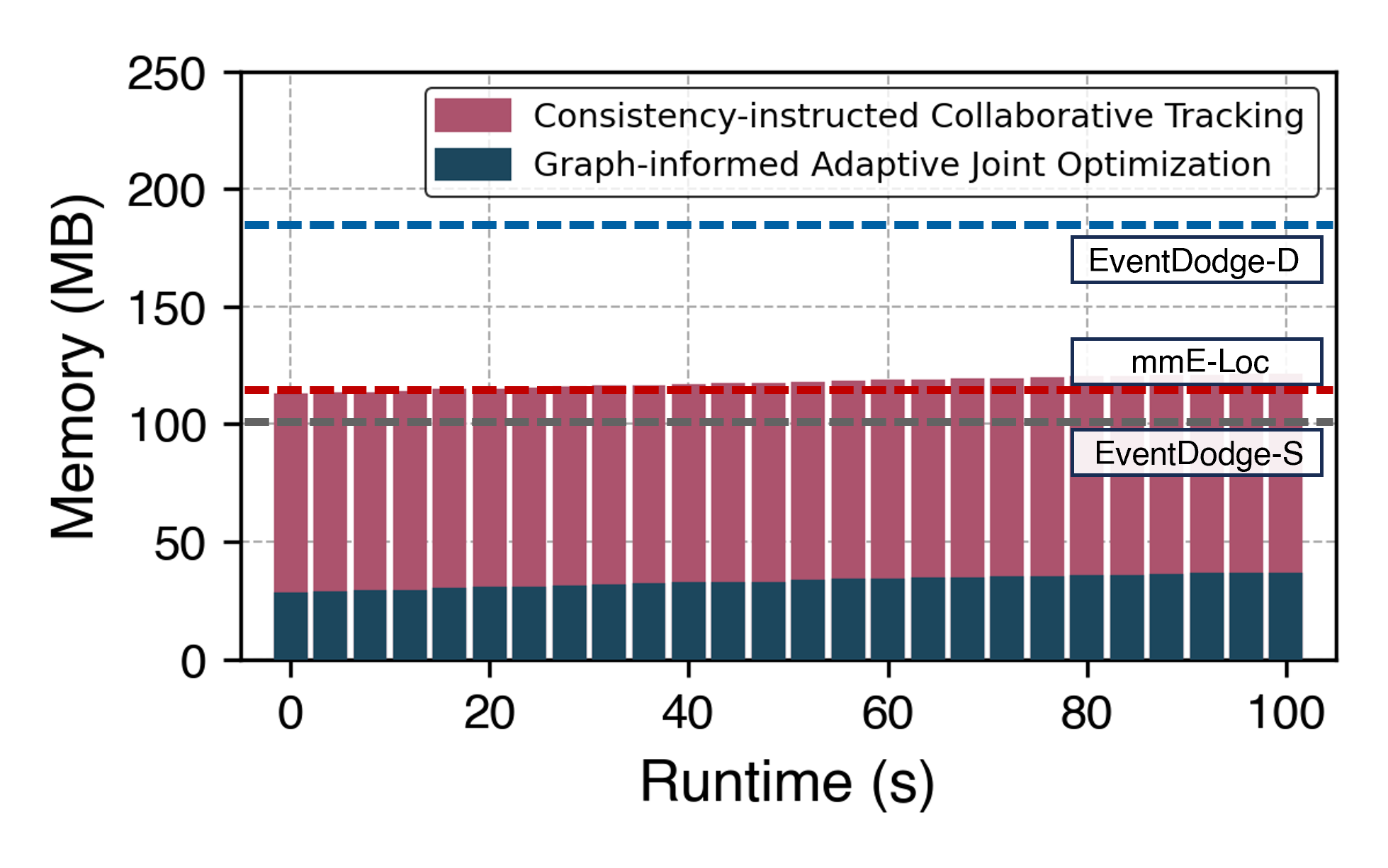}
    \caption{\RR{Memory usage}}
    \label{fig:memory}
  \end{minipage}
  \hfill
  \vspace{-0.4cm}
\end{figure*}
\textbf{Impact of occlusion.}
We validate mmE-Loc's performance with partially occluded drones. 
\TMCrevise{Occlusion is introduced by partially obstructing the drone within the event camera’s field of view, including cases where the drone appears only partially in FoV.}
\fig \ref{fig:occlusion} shows that with 25\% occlusions, mmE-Loc maintains high performance with an average error of 0.094$m$. 
At 50\% occlusion, the average error of mmE-Loc increases to 0.12$m$. 
EventDodge-S and EventDodge-D show larger errors due to incorrect depth estimation caused by occlusion. 
mmE-Loc harnesses the strengths of both modalities in accurately tracking the drone’s location, even under partial occlusion.


\RR{
\textbf{Impact of drone velocity.}
We further evaluate mmE-Loc's performance under different drone velocities in \fig \ref{fig:velocity}. 
The application scenario of mmE-Loc focuses on landing guidance when a delivery drone approaches the landing pad. During this phase, the drone typically reduces its speed to ensure precise and stable landing. Accordingly, we categorize the flight velocities into three levels: slow ($v < 0.5$ m/s), medium ($0.5$ m/s $\leq$ $v$ $<$ $1$ m/s), and rapid ($1$ m/s $\leq$ $v$ $<$ $1.5$ m/s), corresponding to different stages of the drone’s landing process.
Although errors of all methods increased with speed, mmE-Loc maintained an average error of 0.11$m$ even in rapid speed scenarios, outperforming baselines by 44\%, 68.5\%, and 52.4\%.
This benefit stems from the high sampling rates of both the event camera and mmWave radar, which enable more frequent observations of the drone, including both 2D imaging and depth measurements.
}

\vspace{-0.1cm}
\subsection{Ablation study} \label{5.4}
We experimentally analyze the core components of mmE-Loc, \TMCrevise{focusing on both their individual performance and the improvements they contribute to the overall system}.

\textbf{Effectiveness of multi-modal fusion.}
We demonstrate the superiority of fusing radar and event cameras over using each sensor individually. 
\fig \ref{fig:fusion} shows that the fusion-based approach significantly outperforms both radar-only and event camera-only methods in terms of location error. 
mmE-Loc outperforms event-only approach by 63.6\% and exceeds radar-only method by 52.9\%.

\textbf{Contributions of each module.}
We investigate the contributions of \textit{CCT} and \textit{GAJO} to mmE-Loc by gradually integrating them with the event camera into the baseline system (\ie, the radar only-based method) and assessing localization accuracy and end-to-end latency. 
\fig \ref{fig:module} illustrates that without these modules, the baseline method achieves a localization error of 0.229$m$ and latency of 10.19$ms$.
Integrating the event camera with the \textit{CCT} module reduces the localization error to 0.178$m$ and decreases latency to 7.27$ms$.
\TMCrevise{This is because the event camera, together with the \textit{CCT} module, filters out the noisy point cloud generated by the radar.}
Integrating the event camera with \textit{GAJO} further reduces the error to 0.139$m$, although the delay increases due to the absence of an efficient detection mechanism. 
\TMCrevise{This is because the \textit{GAJO} jointly optimize multiple locations to improve accuracy; however, without a robust detection strategy, the system may also optimize the locations of non-drone objects, leading to additional computational overhead.}
Finally, integrating both \textit{CCT} and \textit{GAJO} minimizes both the error and latency.




\RR{
\textbf{Performance of \textit{CCT}.}
We evaluate the filtering performance of the proposed \textit{CCT} module on both mmWave radar and event data in indoor and outdoor environments.
As illustrated in \fig \ref{fig:CCT}, a higher recall indicates that more true drone-triggered data are correctly identified (fewer are missed), while a higher precision reflects that fewer background data are mistakenly classified as drone data (fewer false alarms).
In the indoor scenario, \textit{CCT} achieves recalls above 86\% for mmWave radar and 89\% for the event camera, with corresponding precisions of over 86\% and 82\%. 
The slightly lower filtering rate in indoor conditions can be attributed to two factors:
first, the denoising algorithm may inadvertently remove weak perception signals from the edges of the drone’s propellers; 
second, the complex indoor background, particularly the overhead mesh-like protective structures, can introduce spurious detections for both sensors.
In contrast, under outdoor conditions, the recall remains above 80\% for both sensing types, and the precision exceeds 90\%, demonstrating the robustness of \textit{CCT}. 
Overall, these results arise from two key design advantages of the \textit{CCT} module.
First, it effectively exploits the complementary sensing characteristics of the mmWave radar and event camera, achieving consistent filtering performance across different environments.
Second, the module further incorporates the drone’s intrinsic physical priors, including its periodic micro-motions and characteristic structural patterns, to differentiate genuine drone responses from irrelevant background signals.
In practical terms, this physical knowledge allows the system to recognize and suppress detections that do not exhibit the periodic micro-motions or geometric structure typical of a flying drone.
By leveraging these motion and structural cues, the module effectively filters out spurious responses caused by environmental noise or static reflections.
Consequently, the integration of these physical priors contributes to a consistent improvement in precision for both sensing modalities under both indoor and outdoor conditions, demonstrating the effectiveness of the proposed filtering mechanism.
}

\RR{
\textbf{Performance of \textit{GAJO}.}
We compare the performance of different multi-modal fusion strategies. 
Specifically, we evaluate mmE-Loc against two widely used approaches: the extended Kalman filter (EKF)~\cite{Probabilistic_Robotics} and Graph Optimization (GO) ~\cite{Probabilistic_Robotics}. 
As shown in \fig \ref{fig:GAJO}, mmE-Loc enhances localization performance by over 57.9\% compared to EKF and 47.1\% compared to GO, due to its tightly coupled multi-modal fusion and the incremental optimization method.
We further evaluated the effect of enabling the motion-aware adaptive optimization scheme in mmE-Loc. 
This scheme dynamically adjusts the optimization window based on the drone’s motion state.
Experimental results show that enabling this scheme leads to a 14.9\% increase in localization error, mainly due to the smaller optimization window, which limits the available temporal information for refinement. However, it also yields an 11.5\% reduction in CPU usage, demonstrating that the adaptive mechanism effectively decreases computational overhead.
These findings reveal a clear trade-off between localization accuracy and computational efficiency. 
While a smaller optimization window leads to less accurate but faster updates, it significantly reduces runtime overhead. Consequently, under resource-constrained conditions, such as embedded processors or low-power edge platforms, enabling this scheme can provide a practical balance, maintaining reasonable localization accuracy while improving real-time performance.
}
\vspace{-0.3cm}
\subsection{System efficiency study} \label{5.5}
\RR{
The mmE-Loc distinguishes itself from existing models and learning-based methods due to its low latency and minimal resource overhead. 
\fig \ref{fig:latency} presents the end-to-end latency of the typical localization process, which includes delays from both CCT and GAJO modules.
During drone localization, mmE-Loc’s latency may vary because its motion-aware adaptive optimization scheme optimizes different sets of locations at different times, and the incremental optimization method dynamically decides whether to perform incremental or full optimization at each time step. 
As the size of the location set increases, the optimization time also grows. 
Nevertheless, thanks to the motion-aware adaptive optimization scheme and the incremental optimization method, mmE-Loc maintains low latency overall.
Nonetheless, mmE-Loc's latency remains suitable for use in flight control loops. 
\fig \ref{fig:cpu} and \fig \ref{fig:memory} show that the CPU utilization of mmE-Loc remains below 18\%, while its memory consumption stays under 120 MB. Moreover, the memory usage increases slightly as the size of the location set grows.
To further highlight the efficiency of mmE-Loc, we compare its CPU and memory usage with those of baseline methods (mDrone and milliEye are excluded from this comparison since they rely on GPU acceleration).
As shown in \fig \ref{fig:cpu}, mmE-Loc exhibits an average CPU usage of 7.47\%. The CPU usage of EventDodge-S is about 15\% lower than that of mmE-Loc, primarily because mmE-Loc incorporates mmWave radar for depth estimation, which introduces additional computational overhead while yielding more accurate localization results.
In contrast, EventDodge-D, which employs two event cameras, shows a 125.9\% increase in CPU usage compared to mmE-Loc.
Although multiple event cameras improve localization accuracy, they also introduce significant overhead due to multi-camera target detection and joint position optimization.
As shown in \fig \ref{fig:memory}, the average memory usage of mmE-Loc is 115 MB. EventDodge-S consumes about 13\% less memory, mainly due to the reduced amount of sensor data. In contrast, EventDodge-D exhibits a 60\% higher memory footprint than mmE-Loc, as it processes a larger volume of event camera data and involves more optimization factors.
}

\begin{figure}[t]
    \setlength{\abovecaptionskip}{0.2cm} 
    \setlength{\belowcaptionskip}{-0.5cm}
    \setlength{\subfigcapskip}{-1cm}
    \centering
        \includegraphics[width=0.95\columnwidth]{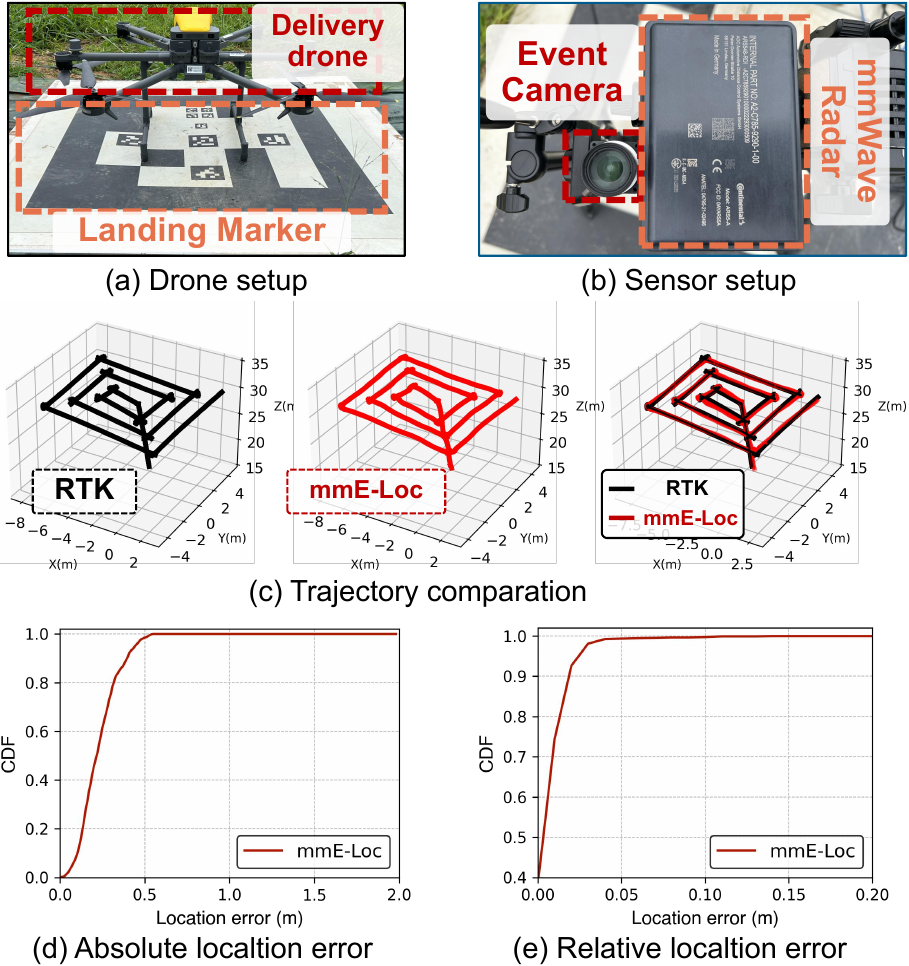}
        \vspace{-0.15cm}
    \caption{\TMCrevise{Case study: delivery drone airport. }}
    \label{airport}
    \vspace{-0.15cm}
\end{figure}


\vspace{-0.2cm}
\TMCrevise{
\section{Case Study}
\subsection{Landing at real-world drone airport}
As shown in \fig \ref{airport}a, to verify the system's usability, we conduct an experiment using a custom drone equipped with six propellers and managed by a PX4 flight controller. 
This drone is developed by a world-class delivery company exploring the feasibility of instant deliveries.
The experiment takes place at a real-world delivery drone airport.
To enable drone localization over a larger area, we employ an ARS548 mmWave radar, as depicted in \fig \ref{airport}b. 
\fig \ref{airport}c, \fig \ref{airport}d, and \fig \ref{airport}e illustrate the localization results as the drone follows a square spiral trajectory at an altitude of 30 $m$.
In the qualitative analysis, the left image in \fig \ref{airport}c shows the localization results from RTK, while the middle image presents the results from mmE-Loc. As illustrated in the right image of \fig \ref{airport}c, mmE-Loc produces smooth trajectories that closely align with those of RTK, demonstrating its potential to complement RTK-based localization.
In the quantitative analysis, the results in \fig \ref{airport}d and \fig \ref{airport}e show that in real-world scenarios, mmE-Loc achieves high localization precision, maintaining a maximum absolute location error error below 0.5$m$ and relative location error error below 0.1$m$. 
mmE-Loc has significant potential as a complementary system to RTK, aiding drone in challenging environments (\eg, urban canyons, where RTK may be compromised by signal blockage).


}
\section{Related work}


\textbf{Drone ground localization.}
Drones have been widely applied in various scenarios.
Several types of work have been proposed to assist drones in localization.
$(i)$ \textit{Satellite-based systems}.
The Global Positioning System (GPS) provides $m$-level accuracy outdoors \cite{wang2023global}, while RTK achieves $cm$-level precision but is costly. 
However, these satellite-based systems struggle in urban canyons \cite{wang2025aerial}. 
$(ii)$ \TMCrevise{\textit{Compact mobile sensor-based systems}.}
To address these issues, various sensor-based techniques are proposed, including camera \cite{ben2022edge}, radar \cite{sie2023batmobility}, LiDAR \cite{wang2025enabling} and acoustic \cite{wang2022micnest} often combined with SLAM (Simultaneous Localization and Mapping) \cite{chen2020h} or deep learning algorithms \cite{zhao2024understanding, zhao2024foes}, aim to improve drone localization. 
However, the limited spatio-temporal resolution of these sensors hinders precise, low-latency localization required for accurate drone landings. 
For instance, cameras and LiDARs typically operate at low frame rates (below 50$Hz$), which may fail to capture rapid motion between frames \cite{jian2023path, zhao2025urbanvideo}.
Acoustic signal-based systems are highly vulnerable to environmental noise, with the presence of nearby humans affecting their stability. 
Visual marker-based systems, which rely on downward-facing cameras mounted on drones, are also sensitive to lighting conditions due to the limited dynamic range of cameras.

Compared to previous methods, mmE-Loc introduces a novel sensor configuration that combines an event camera with mmWave radar, leveraging their ultra-high sampling rates to achieve high-accuracy ground localization for drones with low latency. 
MmE-Loc is resistant to lighting variations thanks to its sensor design: event cameras provide a high dynamic range, while mmWave radar, being based on radio waves, is inherently unaffected by lighting conditions.
It is important to emphasize that mmE-Loc is designed to complement, not replace, existing localization solutions. 
To ensure precise landings, mmE-Loc can be integrated with RTK and visual marker systems, delivering a more reliable localization service.

\textbf{mmWave radar for tracking and localization.}
Millimeter-wave is highly sensitive and more accurate due to its mm-level wavelength.
mmWave radar offers high sensitivity and precision due to its sub-millimeter wavelength \cite{Harlow_2024}. 
Several mmWave radar-based solutions for drone ground localization combine signal intensity methods, but face challenges in accurately tracking the drone's center \cite{qian20203d, asi6040068}. 
This difficulty arises from the drone's large size (\eg, 80 $cm$ across), causing it to appear as a non-uniform blob in radar returns. 
Additionally, these methods often produce unstable results with frequent outliers, as multipath scattering can obscure the main signal and low spatial resolution of radar \cite{Li2024AHH}.
Other approaches leverage deep learning-based methods but require extensive pre-modeling and neural network training for each drone model \cite{zheng2023neuroradar, ALLQUBAYDHI2024100614}. 
These methods tend to struggle with tracking different drone models and perform poorly in environments not represented in training dataset. 
Several solutions integrate visual sensors to assist radar \cite{shuai2021millieye, chadwick2019distant}. 
However, these approaches introduce latency due to exposure times and image processing delays.


\RR{
To address the challenges of accuracy and latency in drone localization, we replace conventional RGB cameras with event cameras and pair them with mmWave radar. 
Event cameras provide high temporal resolution and HDR, offering significant advantages in drone detection tasks.
Recent works have demonstrated their effectiveness: 
\cite{mitrokhin2018event} proposed a time-image representation with feature-less motion compensation for object tracking; 
\cite{kang2024direct} achieved 3D model-based object tracking; 
\cite{chen2019asynchronous} introduced adaptive time-surface-based tracking; 
\cite{da2025new} employed stereo fisheye event cameras for fast drone detection and tracking; 
and \cite{zhang2025evdetmav} presented a generalized drone detection framework from moving event cameras.
These studies collectively demonstrate the strong advantages of event cameras in drone detection, effectively compensating for the limitations of mmWave radar.
Conversely, mmWave radar provides reliable depth information, which complements event-camera perception and further facilitates precise drone localization.
Specifically, by exploiting temporal consistency and the drone’s inherent micro-motions through the \textit{CCT} module, mmE-Loc achieves high-precision tracking without prior knowledge.
The \textit{GAJO} module further enhances ground localization accuracy and reduces latency by leveraging spatial complementarity and motion information.
}

\section{Discussion}\label{8}

\textbf{How does mmE-Loc relate to visual markers?}
Currently, delivery drones utilize visual markers and onboard cameras for self-localization. 
In contrast, mmE-Loc focuses on ground-based drone localization, enabling the landing pad to determine the spatial relationship between the drone and itself for precise adjustments. 
In practice, mmE-Loc operates alongside visual markers to enhance reliability of localization service.

\textbf{Is it feasible to design an onboard sensor system for drone landing localization?}
It's theoretically feasible. 
However, designing an onboard system for drone localization using an event camera and mmWave radar presents several challenges. 
Given the continuous motion of the drone, the system must address:  
$(i)$ motion compensation for event data,  
$(ii)$ reliable feature extraction and matching within the event stream despite its lack of semantic information, and  
$(iii)$ the sparsity of radar measurements and noise induced by specular reflections, diffraction, and multi-path effects.

\textbf{How does network latency impact the system, and how can potential delay-related issues be addressed?}
mmE-Loc is designed to assist ground platforms in locating drones and guiding them to land accurately at designated spots.
Network latency may impact the location update rate for the drone. 
To mitigate this issue, drone airports usually deploy access points near the landing pad and utilize multiple wireless links, including Wi-Fi and cellular networks, to ensure reliable and fast communication.
Additionally, an optics-based communication system could be integrated into the system to ensure a high communication frequency \cite{wang2024towards}.

\RR{
\textbf{Will there be cases that the drone is not fully visible in the camera frame, as mentioned in \cite{springer2022autonomous}?}
Yes, such cases may occur under certain extreme conditions. 
For instance, when the drone descends to a very low altitude (e.g., around 1 m) and is about to land on the pad, it may become not fully visible in the camera frame.
At this stage, mmE-Loc has already guided the drone directly above the pad, and the final descent can be safely completed using the onboard IMU.
Similarly, when the drone is at high altitude and just entering mmE-Loc’s operational range, partial visibility may also occur.
In practice, given the critical importance of flight safety in commercial operations, multiple localization systems, such as visual markers, visual positioning services, and RTK, are used together. 
Thus, even if mmE-Loc does not provide optimal results, stable and safe flight remains fully ensured.
}

\RR{
\textbf{Does mmE-Loc address orientation ambiguities that arise from the use of some fiducial markers, as observed by \cite{springer2022evaluation}?} $(i)$ During visual-marker-assisted landing, our drone did not encounter orientation ambiguities, as it employs a fixed, downward-facing frame camera. Unlike gimbal-mounted cameras, this configuration maintains a constant orientation relative to the drone body, preventing cases where a marker appears identical from different viewing angles.
$(ii)$ Likewise, in mmE-Loc-assisted landing, no orientation ambiguities were observed. Before transmitting localization results, we transform them into the global coordinate system, as the mmE-Loc deployment location is known in advance. Operating within this unified frame, the drone can accurately adjust its position and achieve a stable, precise landing.
}

\RR{
\textbf{Why is the mmWave + event camera configuration preferable despite higher cost and limited field of view?} 
First, our goal is to achieve accurate and low-latency drone localization for safe landings, which only the event camera and mmWave radar can provide. 
Frame cameras and LiDARs suffer from low frame rates, limiting real-time performance. 
Second, as the drone-driven low-altitude economy is projected to reach \$1 trillion by 2040, one event camera + mmWave radar setup per airport is economically acceptable since it serves multiple drones. 
Finally, with the rapid advancement of event-based sensors, the cost is steadily decreasing; for example, Prophesee’s GenX320 is priced around \$300 \cite{GenX320}. 
Regarding the FoV, mmE-Loc focuses on the terminal landing phase, where such coverage is sufficient.
}

\RR{
\textbf{How cost-effective are existing methods, and what scenarios can the new combination uniquely address?}
Current drone localization methods mainly rely on frame cameras, LiDAR, or mmWave radars, each with notable limitations. Frame cameras, such as the Logitech C920E (1080p, 30Hz, \$50), provide sufficient spatial resolution but suffer from long exposure times, low frame rates (typically below 50Hz), and sensitivity to lighting conditions, which reduces localization accuracy in high-speed or low-altitude operations. LiDAR sensors, such as Velodyne VLP-16 (\$4,000) or Livox Mid-40 (\$500), generate dense 3D point clouds but are limited by low scanning frequencies (typically below 20Hz), sparse returns at long ranges, high cost, and heavy onboard processing requirements. mmWave radars, e.g., TI IWR1843 BOOST (\$400), achieve ms-level temporal resolution and reliable radial depth estimation, yet their horizontal and vertical spatial resolution is limited, constraining 3D localization performance.
Our proposed mmE-Loc fuses an event camera with mmWave radar. 
Event cameras provide high-temporal-resolution, HDR-enabled visual sensing, complementing the radar’s radial measurements and limited spatial resolution. 
Both sensors produce sparse outputs focusing on dynamic objects, reducing processing demands while enabling rapid, low-latency tracking of drones, thus supporting faster and more efficient drone operations.
The fusion supports operation under varying illumination, including low-light or nighttime conditions. 
Although the combined system has higher upfront costs than a single sensor, one event + mmWave unit can serve multiple drones in shared operational areas. 
Given the projected \$1 trillion low-altitude drone economy by 2040 and decreasing event camera costs (e.g., Prophesee GenX320, \$300), this investment is economically justified. 
The combination offers unique advantages of low-latency, high-accuracy, sparse representation, and suitability for high-speed operations, unattainable by conventional camera, LiDAR, or radar-only methods.
}

\RR{
\textbf{How does mmE-Loc handle multi-drone discrimination and tracking?}
mmE-Loc distinguishes and tracks multiple drones by exploiting their distinct periodic micro-motion patterns and the corresponding signatures captured by the mmWave radar and event camera. 
Specifically, drones exhibit unique periodic motion characteristics when performing different maneuvers (e.g., ascending, descending, or lateral motion).
Meanwhile, the bounding box size and position observed in the event camera (representing spatial cues), as well as the range measurements from the mmWave radar (representing distance cues), also differ among drones.
By associating each drone with a unique combination of periodic micro-motion pattern, bounding box characteristics, and range information, mmE-Loc can reliably identify and track individual drones.
In real-world drone delivery scenarios, a landing pad typically allows only one drone to land at a time, while others hover nearby awaiting clearance. In such cases, mmE-Loc utilizes periodic micro-motion patterns to distinguish between the landing drone and hovering drones, and further discriminates among multiple hovering drones based on their bounding box positions and sizes in the event camera and their corresponding range measurements in the mmWave radar.
}
\vspace{-0.2cm}
\section{Conclusion}\label{7}

\TMCrevise{
This paper presents a novel sensor configuration that combines an event camera with mmWave radar, synchronizing their ultra-high sampling frequencies to enable mmE-Loc, a ground localization system designed for drone landings, achieving cm-level accuracy and ms-level latency.
mmE-Loc comprises two key components:
$(i)$ consistency-instructed collaborative tracking (CCT) module, which exploits cross-modal temporal consistency along with the drone's physical knowledge of periodic micro-motions and structure to enable precise drone detection; and
$(ii)$ graph-informed adaptive joint optimization (GAJO) module, which enhances localization accuracy and reduces latency by leveraging cross-modal spatial complementarity. 
Extensive evaluations across diverse scenarios validate effectiveness of mmE-Loc.
}



\bibliographystyle{unsrt}
\bibliography{reference}

\end{document}